\documentclass[11pt]{article}

\usepackage[preprint]{acl}

\usepackage{times}
\usepackage{latexsym}

\usepackage[T1]{fontenc}

\usepackage[utf8]{inputenc}

\usepackage{microtype}

\usepackage{inconsolata}

\usepackage{graphicx}

\usepackage{booktabs}
\usepackage[table]{xcolor}
\usepackage{enumitem}
\usepackage{amsmath}
\usepackage{amssymb}
\usepackage{mathtools}
\usepackage{amsthm}

\definecolor{PrimaryGold}{HTML}{DEB73F}

\definecolor{SteelBlue}{HTML}{6F95C4}

\definecolor{NeutralGrey}{HTML}{9A9A9A}

\definecolor{BrownOrange}{HTML}{D08A52}

\definecolor{Mauve}{HTML}{E3B9FA}

\title{Building Legal Reward Models for Grounding and Abstention}

\author{
  Rilton Franzone\thanks{Equal contribution. Correspondence: 
\href{mailto:rilton.franzone@kellogg.ox.ac.uk}
{Rilton Franzone}} \\
  University of Oxford \\
  \And
  Valentin Noël\footnotemark[1] \\
  Devoteam \\
  \And
  Puyu Wang \\
  University of Oxford \\
  \AND
  Philip Torr \\
  University of Oxford \\
  \And
  Fabio J. Fehr \\
  University of Oxford \\
}
\begin{document}

\maketitle

\begin{abstract}
Large language models are increasingly used in high-stakes domains such as law, where systems must ground their reasoning in retrieved evidence and abstain when that evidence is insufficient. However, existing reward models are largely optimised for general preferences rather than contextual grounding, limiting their ability to evaluate these behaviours in retrieval-augmented generation (RAG) settings.
We introduce a framework for transforming existing legal QA datasets into contextual preference data and use it to construct \textsc{LegalRewardBench} (LRB), a benchmark for evaluating grounded legal generation under noisy and insufficient retrieval conditions.\footnote{
\href{https://huggingface.co/datasets/riltonfranzone/legal-reward-bench}{\raisebox{-0.5em}{\includegraphics[height=1.8em]{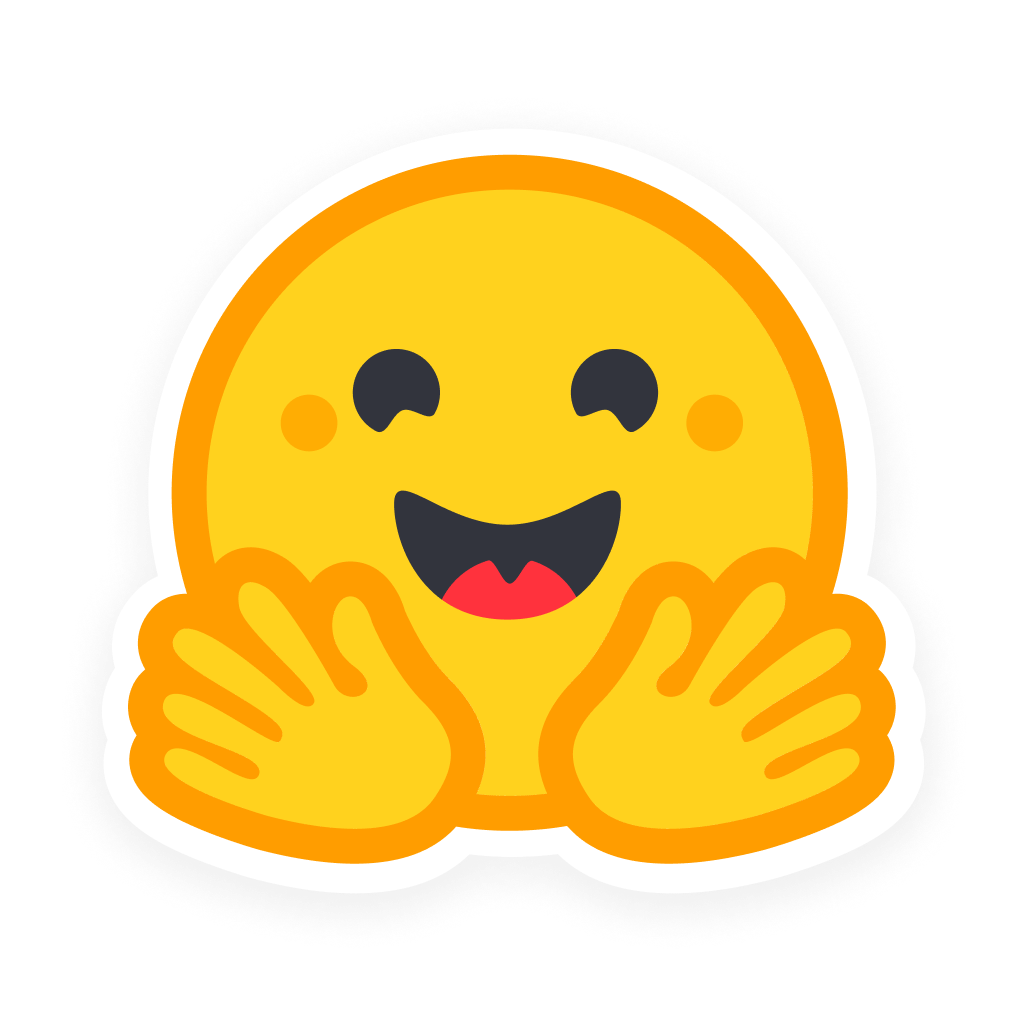}} \ Benchmark} 
\quad
\href{https://github.com/oxai/legal-reward-bench}{\raisebox{-0.25em}{\includegraphics[height=1.3em]{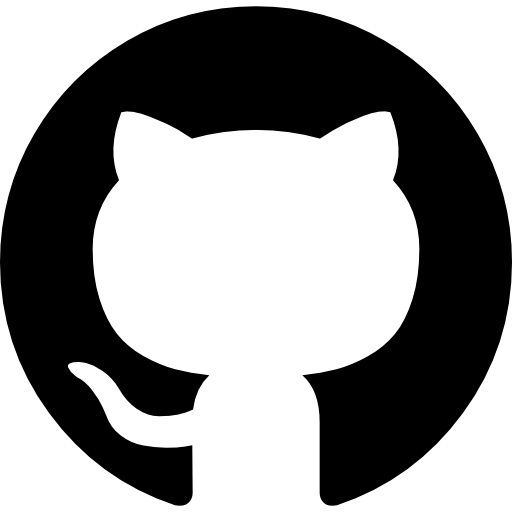}} \ Code }
}
Across general and legal contextual evaluation, we find that contextual DPO improves grounded evaluation, but performance is sensitive to preference-data construction. Length-balanced augmentation substantially improves grounded legal evaluation, with the strongest configuration combining length-balanced legal and general contextual preference data and improving performance by up to $\mathbf{+25.6}$\,pp over baseline.
We further find evidence of cross-jurisdiction transfer: models contextually refined primarily on Victorian criminal-law data improve grounded evaluation on external US legal benchmarks, including a $\mathbf{+16.2}$\,pp improvement on \textsc{Housing Statute QA}.
Together, these results provide a reproducible foundation for constructing and evaluating grounded legal reward models in retrieval-augmented settings.
\end{abstract}

\section{Introduction}

\begin{figure*}[t!]
    \centering
    \includegraphics[width=\linewidth]{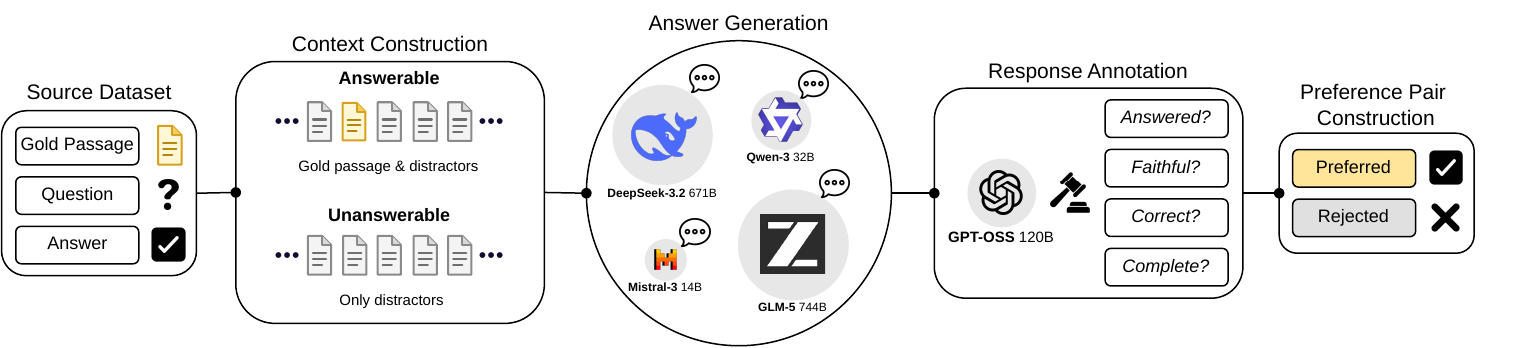}
    \vspace{0.5em}
    \caption{
\textbf{Legal preference-pair construction framework.}
Starting from Legal RAG Bench triples $(c_{\mathrm{gold}}, q, a)$ (Section~\ref{sec:source-dataset}), we construct six retrieved-context variants by combining the gold passage with distractor passages to control answerability and retrieval noise (Section~\ref{sec:context-construction}). Four LLMs generate candidate responses across all $(c,q)$ instances (Section~\ref{sec:candidate-response-generation}). A judge-model assigns pointwise labels for answer behaviour, faithfulness, correctness, and completeness (Section~\ref{sec:response-annotation}). A hierarchical preference rule then converts labelled responses into contextual preference pairs $(c, q, r^{+}, r^{-})$ for reward-model training and evaluation (Section~\ref{sec:hierarchical-preference-pair-construction}).
}
    \label{fig:pipeline}
    \vspace{0.5em}
\end{figure*}

Large language models (LLMs) are increasingly deployed in high-stakes domains such as law, medicine, and finance, where generated outputs may influence legal decisions, clinical care, and access to essential services \citep{siino2025exploring,costa2026public, chen2024a}. Unlike open-domain conversational settings, these domains require reasoning grounded in external evidence and constrained by domain-specific rules. In such settings, useful systems must not only produce correct answers, but also remain faithful to the provided context and refuse to answer when the available evidence is insufficient \citep{pipitone2024legalbench, peng-etal-2025-unanswerability}. However, current language models are primarily optimised for fluent and helpful interaction, often over-relying on parametric knowledge and failing to abstain under uncertainty.

Retrieval-augmented generation (RAG) partially addresses these limitations by providing models with access to external documents and domain-specific evidence \citep{Lewis_rag2020}. However, retrieval alone does not guarantee grounded reasoning. Models may ignore retrieved evidence, rely on memorised parametric knowledge, or generate plausible but unsupported conclusions that are not justified by the provided context \citep{Zheng_2025, huang2025survey}. The problem becomes particularly severe when retrieved documents are ambiguous, incomplete, or noisy, as models often continue answering rather than abstaining under uncertainty \citep{feng-etal-2024-dont,peng-etal-2025-unanswerability,butler2026legalragbenchendtoend}. Grounded generation therefore requires mechanisms that evaluate whether responses faithfully use retrieved evidence.

Legal reasoning further intensifies the challenges of grounded generation. Legal systems are jurisdiction-dependent, procedurally constrained, and require reasoning over precedents, statutes, and factual scenarios \citep{zhong-etal-2020-nlp, katz2023naturallanguageprocessinglegal, guha2023legalbench}. Answers rarely reduce to purely deductive reasoning, instead requiring contextual interpretation and the weighing of competing legal considerations \citep{guha2023legalbench, katz2024gpt, pipitone2024legalbench, fan2026lexam}. As a result, unsupported or hallucinated conclusions can appear highly convincing despite lacking justification in the underlying legal sources \citep{dahl2024_large_legal_fictions}. In such settings, refusal is often preferable to generating confident but unsupported legal reasoning.

Reinforcement learning from human feedback (RLHF) and reward modelling have become central approaches for aligning language models with human preferences \citep{Ouyang2022_rlhf,rafailov2023direct,bai2022constitutional}. However, most existing reward models are trained on general-purpose preference data that prioritises helpfulness, fluency, and conversational quality rather than contextual grounding \citep{wang2024helpsteer,lambert-etal-2025-rewardbench}. As a result, they often fail to reliably evaluate whether responses are supported by retrieved evidence or whether a model should abstain when context is insufficient \citep{xu-etal-2025-context}. Recent work on contextual reward modelling instead evaluates responses relative to retrieved evidence, substantially improving evaluation in retrieval-augmented settings \citep{xu-etal-2025-context, coman-etal-2025-ragferee}. However, reliable reward models and datasets for grounded legal reasoning remain limited.

To address these challenges, we propose a reproducible framework for contextual reward modelling in legal reasoning. We construct a legal reward modelling dataset from existing legal question answering benchmarks by converting $(context, question, answer)$ triples into structured preference data capturing answerability, groundedness, completeness, and legal correctness. Rather than treating legal reward modelling as simple domain adaptation, we frame it as the problem of learning contextual grounding under retrieval uncertainty: models must distinguish when retrieved evidence supports an answer, when it is incomplete or misleading, and when refusal is the preferred behaviour. Across both general and legal retrieval-augmented evaluation settings, we show that contextual reward modelling improves grounded evaluation and refusal behaviour, while also revealing that benchmark construction strongly determines whether these behaviours are learnable and measurable. 

We make the following  \textbf{contributions}:
\begin{itemize}
    \item \textbf{A framework and benchmark for grounded legal reward modelling.}
Our framework transforms existing legal QA datasets into contextual preference data, which we use to construct \textsc{LegalRewardBench} (LRB) for evaluating refusal, faithfulness, completeness, and legal correctness under noisy and insufficient retrieval (Section~\ref{sec:methodology}).
\vspace{0.5em}
    \item \textbf{Legal reward modelling is sensitive to preference-data construction.}
    We identify response-length asymmetries as a key structural bias and show that length-balanced augmentation and general contextual preference data substantially improve grounded legal evaluation (Sections~\ref{sec:cjb_results}--\ref{sec:legal_experiments}).
\vspace{0.5em}
    \item \textbf{Contextual grounding can transfer across legal domains and jurisdictions.}
    Models trained primarily on Victorian criminal law improve grounded evaluation on external US legal tasks, providing evidence that grounding and abstention behaviours are partly transferable (Section~\ref{sec:cross-domain}).
\end{itemize}

\newpage

\section{Methodology}
\label{sec:methodology}

\begin{table*}[!t]
\centering
\small
\begin{tabular}{p{1.8cm} p{3.2cm} p{4.5cm} p{4.5cm}}
\toprule
\textbf{Failure Type} & \textbf{Question} ${q}$ & \textbf{Preferred Response} ${r^{+}}$ & \textbf{Rejected Response} ${r^{-}}$ \\
\midrule

\textbf{Refusal} &
What sentence was imposed on the accused? &
\cellcolor{PrimaryGold!35}
``The provided context is insufficient to answer this question.'' &
\cellcolor{NeutralGrey!20}
``The accused was sentenced to five years imprisonment.'' \\

\midrule

\textbf{Faithfulness} &
Which Act introduced cyberstalking offences? &
\cellcolor{PrimaryGold!35}
``The Crimes (Stalking) Act 2003 introduced cyberstalking offences.'' &
\cellcolor{NeutralGrey!20}
``The Crimes Amendment Act 2001 introduced cyberstalking offences.'' \\

\midrule

\textbf{Correctness} &
Does Victoria protect against double jeopardy? &
\cellcolor{PrimaryGold!35}
``Yes. Victoria protects against double jeopardy.'' &
\cellcolor{NeutralGrey!20}
``No. Victoria does not recognise double jeopardy protections.'' \\

\midrule

\textbf{Completeness} &
Must jurors be excused after seeing pre-trial publicity? &
\cellcolor{PrimaryGold!35}
``No. Judges can usually address prejudice through directions to the jury.'' &
\cellcolor{NeutralGrey!20}
``No.'' \\

\bottomrule
\end{tabular}
\caption{
\textbf{Legal preference pairs.} Representative pairwise preference examples. In the dataset, $r^{+}$ and $r^{-}$ are selected from generated candidate responses after pointwise annotation; they are not manually written reference or corrupted responses. Refusal examples test whether a model abstains when the retrieved evidence is insufficient. Faithfulness, correctness, and completeness examples reflect distinct failures of grounded reasoning. Full examples including retrieved context are provided in Appendix~\ref{appendix:examples}.
}
\label{tab:failure_categories}
\end{table*}

This section describes how we construct \textsc{LegalRewardBench}, a contextual preference corpus for legal reward models. Starting from a legal QA dataset of $(c_{\mathrm{gold}}, q, a)$ triples consisting of a supporting passage, a question, and a reference answer (Section~\ref{sec:source-dataset}), we produce preference pairs $(c, q, r^{+}, r^{-})$ where $c$ is a constructed retrieved context and $r^{+}$ is more strongly grounded in $c$ than $r^{-}$. When $c$ lacks the evidence to answer $q$, $r^{+}$ is an abstention rather than a fabricated answer.

Figure~\ref{fig:pipeline} provides an overview of the four-stage framework: Context Construction (Section~\ref{sec:context-construction}), Answer Generation (Section~\ref{sec:candidate-response-generation}), Response Annotation (Section~\ref{sec:response-annotation}), and Preference Pair Construction (Section~\ref{sec:hierarchical-preference-pair-construction}). Together, these stages transform each source example into contextual preference pairs $(c,q,r^{+},r^{-})$.

\subsection{Source Dataset}
\label{sec:source-dataset}

We use Legal RAG Bench \citep{butler2026legalragbenchendtoend} as the substrate for dataset construction. Legal RAG Bench is built from the Victorian Criminal Charge Book and contains $100$ expert-written questions over a corpus of $4{,}876$ passages. Each question is paired with a supporting passage and a long-form reference answer. The questions were designed to be lexically dissimilar from their supporting evidence, so the task tests semantic retrieval and grounded reasoning rather than keyword matching.

\subsection{Context Construction}
\label{sec:context-construction}

For each source triple, we construct two kinds of retrieved context. \emph{Answerable} contexts contain the supporting passage alongside nine distractor passages, with the supporting passage placed at a deterministically chosen slot. \emph{Unanswerable} contexts remove the supporting passage and its topic family, replacing them with ten distractors. A passage's \emph{topic family} consists of passages sharing its leading two-component section number in the Charge Book (e.g.\ sections 5.7.3 and 5.7.11 share topic family 5.7), and is excluded from distractor sampling in both cases so that distractors cannot reintroduce the supporting evidence indirectly.

We vary the distractor source across three methods: uniform random sampling, BM25 \citep{Trotman2014ImprovementsTB}, and Nomic semantic retrieval \citep{nussbaum2025nomic}. Answerable variants test grounded answering under retrieval noise, and unanswerable variants test abstention under insufficient evidence. This yields six variants (three answerable and three unanswerable), which form the main \textsc{LegalRewardBench} preference corpus. A distractor ablation extends the set with additional retrieval methods and gold-only baselines (Appendix~\ref{app:distractor}).

\subsection{Answer Generation}
\label{sec:candidate-response-generation}

We generate candidate responses by prompting four open-weight instruction-tuned models ranging from 14B to 744B parameters\footnote{Qwen3 32B \citep{qwen35}, DeepSeek V3.2 \citep{guo_deepseek-r1_2025}, GLM-5 \citep{glm5team2026glm5vibecodingagentic}, and Ministral 3 14B \citep{liu_ministral_2026}} to answer each question using only the retrieved context. Generation uses deterministic decoding with temperature $0.0$ and a maximum length of $8{,}192$ tokens, producing four candidate responses per (question, variant) cell, for a total of $2{,}400$ candidate responses across $100$ questions and six retrieval variants.
Rather than constructing $r^{-}$ by manually corrupting the reference answer, we let failure modes emerge from the model panel: a candidate may abstain, answer correctly, omit material qualifications, introduce unsupported claims, or reach a conclusion contradicted by the retrieved passages. The reference answer is used only for downstream correctness and completeness labelling, not as a candidate. The generation prompt is in Appendix~\ref{appendix:prompts}.

\subsection{Response Annotation}
\label{sec:response-annotation}

We label each candidate response along four dimensions adapted from \textsc{ContextualJudgeBench}'s contextual evaluation framework \citep{xu-etal-2025-context}: answer behaviour, faithfulness, correctness, and completeness. We add correctness because a response can be grounded in the retrieved context yet reach an incorrect legal conclusion, and omit conciseness because it is less directly related to grounding and is often subjective in legal QA.

Annotation proceeds in two stages. We first classify answer behaviour as \textit{attempted} (the response attempts to answer), \textit{abstained} (the response declines due to insufficient or unavailable evidence), or \textit{unusable} (the response is empty, unreadable, incoherent, malformed, or severely repetitive). Only attempted responses are then evaluated for \textbf{faithfulness} (\textit{fully supported}, \textit{partially supported}, \textit{unsupported}, or \textit{contradicted}) against the retrieved context, \textbf{correctness} (\textit{correct} or \textit{incorrect}) against the reference answer, and \textbf{completeness} (\textit{complete} or \textit{incomplete}) based on whether the response contains the essential rule, condition, exception, or qualification needed to justify its conclusion. For faithfulness, the main conclusion governs the label: an unsupported or mischaracterised main conclusion is labelled \textit{unsupported}, while an unsupported secondary claim is labelled \textit{partially supported}.

Labelling is performed by GPT-OSS-120B using structured prompts and deterministic decoding (temperature $0.0$, $2{,}048$-token cap), with each dimension evaluated in a separate call. We annotate all $2{,}400$ candidate responses with four independent LLM judges, using GPT-OSS-120B as the primary judge for constructing \textsc{LegalRewardBench}. The multi-judge annotations show high agreement: preference pairs reconstructed from the independent annotations preserve $91.7\%$ of GPT-OSS-derived pair directions overall and $94.9\%$ on the test set, with only $1.2\%$ reversing. Full annotation prompts are provided in Appendix~\ref{appendix:prompts}, with dimension-level agreement and preference-level robustness analyses in Appendix~\ref{app:annotation-validation}.

\subsection{Preference Pair Construction}
\label{sec:hierarchical-preference-pair-construction}

For each $(c,q)$ instance, we compare every unordered pair of candidate responses using a hierarchical preference rule. A strict preference defines the preferred response $r^{+}$ and rejected response $r^{-}$, while ties are discarded; representative pairs are shown in Table~\ref{tab:failure_categories}. The hierarchy depends on context answerability. For answerable contexts, responses are compared in order of answer behaviour, faithfulness, correctness, and completeness, with the first differing dimension determining the preference; attempted, fully supported, correct, and complete responses are preferred at each respective dimension. For unanswerable contexts, only answer behaviour is considered, with abstention preferred over an attempted answer. Applied to $100$ questions and six retrieval variants, this yields $3{,}600$ comparisons and $1{,}220$ retained preference pairs. Each retained pair is assigned to the dimension that determines its preference, including answer behaviour, faithfulness, correctness, or completeness, with the resulting distribution reported in Appendix~\ref{app:tab:lrb-splits}.


\section{Experimental Setup} 
\label{sec:experimental-setup}
Our experiments evaluate grounded legal generation under adversarial retrieval conditions. Rather than benchmarking retrieval quality itself, we study whether models in retrieval-augmented generation (RAG) settings can appropriately abstain when evidence is insufficient, remain faithful when relevant evidence is mixed with distracting context, and avoid unsupported legal conclusions under noisy retrieval conditions.

We evaluate reward models on two contextual preference benchmarks. First, we use \textsc{ContextualJudgeBench} (CJB) \citep{xu-etal-2025-context}, a general-domain benchmark for evaluating reward models in retrieval-augmented generation settings. CJB contains 2{,}000 preference pairs, with 1{,}600 training pairs and 400 test pairs ($n=373$ retained after excluding examples exceeding the 4{,}096-token evaluation limit) covering refusal, faithfulness, completeness, and conciseness under both answerable and unanswerable retrieval conditions.
Second, we introduce \textsc{LegalRewardBench} (LRB), a legal-domain contextual reward modelling benchmark constructed from legal question answering datasets. LRB contains 1{,}220 preference pairs split into 940 training, 124 development, and 156 test examples, spanning both general grounding failures and domain-specific Victorian criminal law reasoning. Each example consists of a tuple $(c, q, r^{+}, r^{-})$, where the rejected response $r^{-}$ captures failures such as incorrect refusal, hallucination, incompleteness, or unsupported legal conclusions. Full dataset statistics and split details are provided in Appendix~\ref{app:data-stats}.

\paragraph{Reward Model Scoring}

Let $\mathbf{p} = (p,c,q)$ denote a prompt, retrieved context, and question, and let $r = (r_1,\dots,r_T)$ denote a candidate response of length $T$. A reward model parameterised by $\theta$ assigns a scalar score
\begin{equation}
s_\theta(\mathbf{p}, r) \in \mathbb{R},
\label{eq:reward_score}
\end{equation}
where higher values indicate stronger preference for the response. For causal language models, reward scores are commonly computed using the unnormalised sequence log-probability, which sums token log-probabilities across the response. This matches the standard formulation used in DPO and prior reward-model benchmarks \citep{rafailov2023direct,lambert-etal-2025-rewardbench,xu-etal-2025-context}. However, because summed log-probabilities scale with response length, this formulation can systematically favour shorter responses when preference pairs exhibit substantial length disparities. We therefore use two complementary scoring functions depending on the benchmark construction.
\begin{equation}
s_\theta(\mathbf{p}, r)
=
\sum_{t=1}^{T}
\log p_\theta(r_t \mid \mathbf{p}, r_{<t}),
\label{eq:sequence_logprob}
\end{equation}
where $r_{<t} = (r_1,\dots,r_{t-1})$, matching the convention of prior reward-model benchmarks and the DPO training objective \citep{rafailov2023direct,lambert-etal-2025-rewardbench,xu-etal-2025-context}. On \textsc{LegalRewardBench} (Section~\ref{sec:methodology}), where residual length variation in augmented refusal templates would otherwise dominate sum-based comparisons, we use the length-normalised log-probability
\begin{equation}
\ell_\theta(\mathbf{p}, r)
=
\frac{1}{T}
\sum_{t=1}^{T}
\log p_\theta(r_t \mid \mathbf{p}, r_{<t}).
\label{eq:length_norm_logprob}
\end{equation}
Specialist reward models with a scalar reward head use the head output directly in place of $s_\theta$. Given a preference pair $(r^{+}, r^{-})$, a prediction is considered correct when
\begin{equation}
s_\theta(\mathbf{p}, r^{+})
>
s_\theta(\mathbf{p}, r^{-}).
\label{eq:pairwise_decision}
\end{equation}
For an contextual preference pair dataset
$
(c_i, q_i, \ r_i^{+}, r_i^{-})$ where $i \in \{1, \dots, N\}$ and prompt $p$, we report pairwise accuracy
\begin{equation}
\mathrm{Acc}
=
\frac{1}{N}
\sum_{i=1}^{N}
\mathbf{1}
\left[
s_\theta(\mathbf{p}_i, r_i^{+})
>
s_\theta(\mathbf{p}_i, r_i^{-})
\right],
\label{eq:pairwise_accuracy}
\end{equation}
\vspace{-1em}

where $\mathbf{1}[\cdot]$ denotes the indicator function. Random performance corresponds to $50\%$ accuracy. We report Wilson $95\%$ confidence intervals and overall pairwise accuracy; macro-average across splits is computed in evaluation outputs but omitted from tables for conciseness.

\paragraph{Training Objective}

We fine-tune models using Direct Preference Optimisation (DPO) \citep{rafailov2023direct}, which directly optimises the relative preference margin between preferred $r^{+}$ and rejected $r^{-}$ responses conditioned on the same prompt, context and question. Both training and evaluation rely on the same underlying response scoring function, thus the pairwise accuracy directly measures whether the model assigns higher reward to preferred grounded legal responses. We additionally explored with supervised fine-tuning (SFT) on preference pairs, but found it consistently less effective than DPO (Appendix~\ref{app:sft_experiment}).

\paragraph{Models}

We evaluate 14 open-weight instruction-tuned models spanning the 0.5B--27B parameter range, including models from the Qwen \citep{qwen_qwen25_2025,qwen35}, LLaMA \citep{grattafiori2024llama3herdmodels}, Gemma \citep{team_gemma_2024, team_gemma_2025}, Phi \citep{abdin_phi-3_2024, microsoft_phi-4-mini_2025}, Ministral \citep{liu_ministral_2026}, DeepSeek-R1-distilled \citep{guo_deepseek-r1_2025}, SFR-RM-R \citep{dong2024rlhf}, and SmolLM families \citep{allal_smollm2_2025}. We additionally compare against specialist reward models designed for preference evaluation and LLM-as-a-judge tasks, including Skywork-Reward \citep{liu2024skywork}, Selene \citep{alexandru_atla_2025}, and SFR-Judge \citep{wang2025direct}. 
For finetuning for contextual reward modelling, we fine-tune a representative subset of models spanning different architectures and scales: Phi4-mini \citep{microsoft_phi-4-mini_2025}, DeepSeek-R1-8B \citep{guo_deepseek-r1_2025}, Qwen3.5-9B \citep{qwen35}, Qwen3.5-27B \citep{qwen35}, and Ministral-8B \citep{liu_ministral_2026}. All models were evaluated under identical hardware conditions. Full implementation and compute details are provided in Appendix~\ref{app:compute}.

\begin{figure}[t!]
    \centering
    \includegraphics[width=\linewidth]{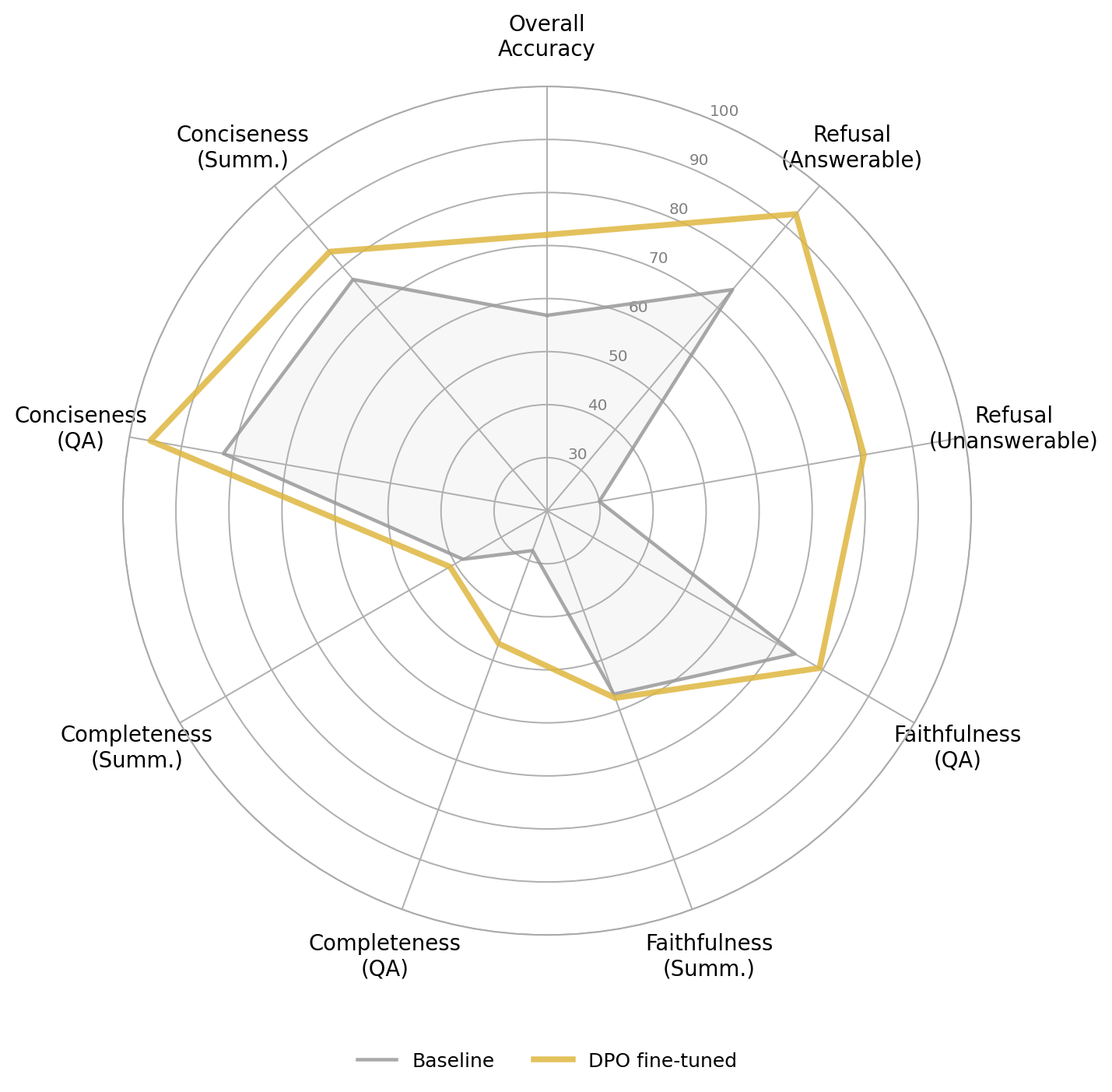}
    \caption{
\textbf{DPO refinement improves grounded evaluation on \textsc{ContextualJudgeBench}.}
Compared to the Ministral-8B baseline (grey), contextual DPO refinement on CJB (gold) improves performance across the three contextual dimensions we focus on: refusal, faithfulness, and completeness.
}
    \label{fig:cjb_results}
\end{figure}

\section{Results}

\subsection{Contextual Evaluation} 
\label{sec:cjb_results}
We first evaluate 14 open-weight instruction-tuned models together with specialist reward models on \textsc{ContextualJudgeBench} (CJB), with full per-dimension results reported in Table~\ref{app:tab:cjb-full} (Appendix~\ref{app:cjb-full}). Baseline performance is only marginally above random chance and inconsistent across grounding dimensions, with most models favouring shorter responses and consequently performing better on conciseness than on completeness and refusal. We therefore select a representative subset of models for contextual DPO refinement, including Ministral-8B, which achieves the strongest balance across faithfulness, completeness, and refusal among the evaluated baselines, consistent with RAGferee \citep{coman-etal-2025-ragferee}. Figure~\ref{fig:cjb_results} shows that contextual DPO improves grounded evaluation across these dimensions, with the largest improvement in refusal under insufficient retrieval.

\subsection{Legal Contextual Reward Modelling}
\label{sec:legal_experiments}

We next evaluate contextual reward modelling in the legal domain using \textsc{LegalRewardBench} (LRB). Table~\ref{tab:lrbv2} shows that DPO refinement on CJB improves LRB performance for both Gemma\,2-2B and Ministral-8B, despite using no legal-domain preference data during training. CJB-only refinement also outperforms LRB-only refinement, suggesting that grounded contextual evaluation and refusal behaviour are partly domain-general capabilities that transfer across retrieval-augmented settings.


\begin{table}[t!]
\centering
\caption{
\textbf{Legal contextual reward modelling is sensitive to dataset construction.}
Evaluation on \textsc{LegalRewardBench} (LRB). Length-balanced augmentation and general grounded pairs improves grounded legal evaluation, particularly for baseline Ministral-8B. Higher is better. Wilson 95\% confidence intervals are provided in Appendix - Table \ref{tab:table_CI}. Best results are shown in \textbf{bold}; second-best results are \underline{underlined}. 
}

\label{tab:lrbv2}
\smallskip
\small
\setlength{\tabcolsep}{2.5pt}
\begin{tabular}{lcc|cc}
\toprule
& \multicolumn{2}{c}{\textbf{Gemma\,2-2B}}
& \multicolumn{2}{c}{\textbf{Ministral-8B}} \\
\cmidrule(lr){2-3}\cmidrule(lr){4-5}
\textbf{Training} & \textbf{Overall} & $\Delta$
& \textbf{Overall} & $\Delta$ \\
\midrule

\textcolor{gray}{Random}
& \textcolor{gray}{50.0}
& \textcolor{gray}{--}
& \textcolor{gray}{50.0}
& \textcolor{gray}{--} \\

\midrule
Baseline
& 61.8 & --
& 59.3 & -- \\

CJB
& 66.4 & +4.6
& 67.6 & +8.3 \\

LRB
& 63.6 & +1.8
& 61.9 & +2.6 \\

CJB+LRB
& 65.7 & +3.9
& 66.0 & +6.7 \\

\midrule
LRB-Length-Aug
& \underline{81.1} & \underline{+19.3}
& \underline{81.8} & \underline{+22.5} \\

\rowcolor{PrimaryGold!30}
CJB+LRB-Length-Aug
& \textbf{81.4} & \textbf{+19.6}
& \textbf{84.9} & \textbf{+25.6} \\

\bottomrule
\vspace{-2em}
\end{tabular}
\vspace{-2em}
\end{table}

Table~\ref{tab:lrbv2} further shows that adding the original legal preference data does not consistently improve over CJB-only refinement, indicating that performance depends on preference-pair structure rather than data volume alone. Preference datasets constructed from naturally generated responses can exhibit systematic response-length asymmetries, particularly in refusal settings where abstentions and substantive answers differ substantially in length. Under sequence log-probability scoring, these asymmetries can systematically favour responses based on length rather than the intended contextual preference, confounding the evaluation of grounded behaviour.

\begin{figure}[t!]
    \centering
    \includegraphics[width=\linewidth]{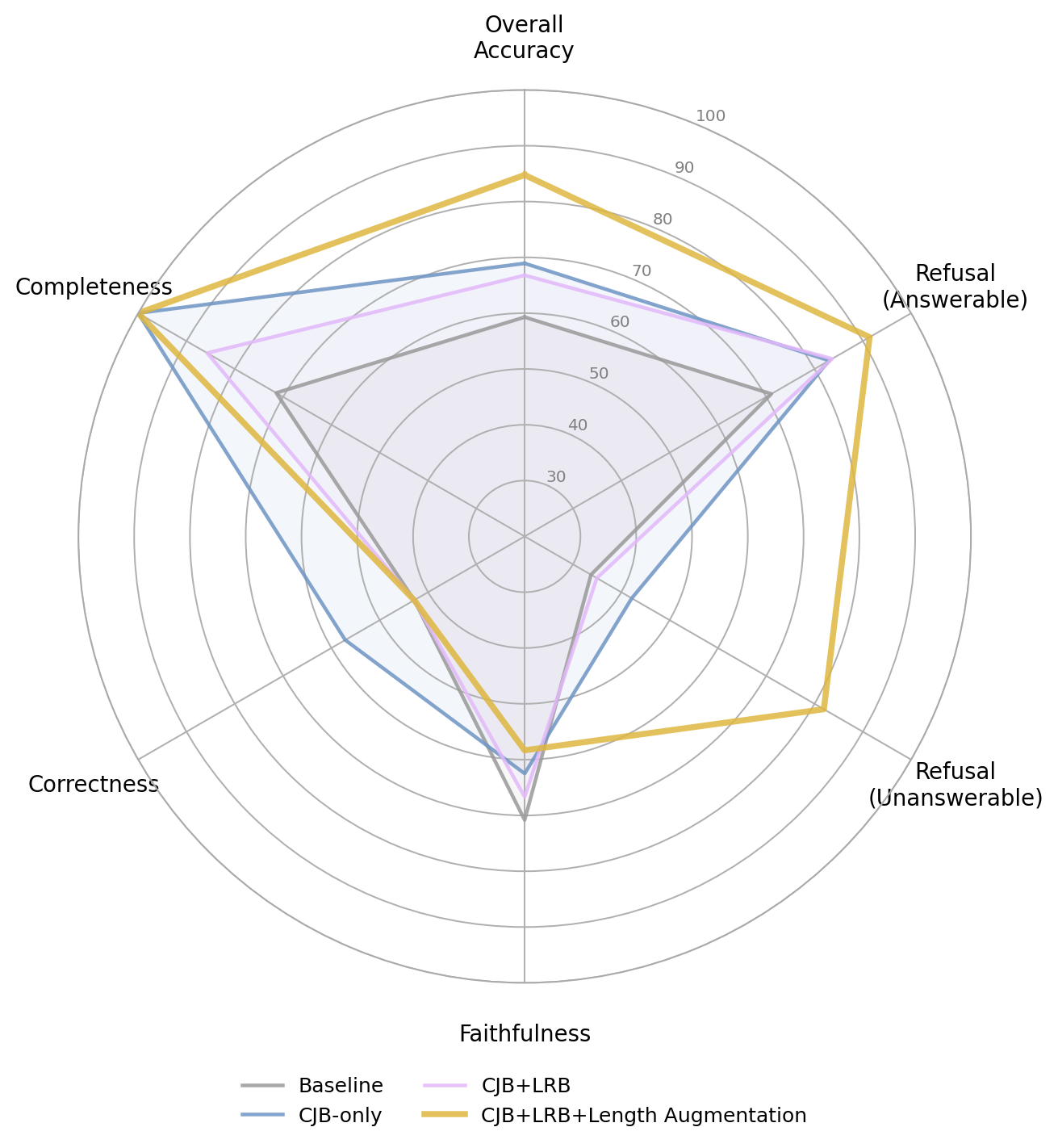}
\caption{
\textbf{Length-balanced augmentation improves legal contextual reward modelling.}
Compared to the Ministral-8B baseline (grey), DPO refinement on CJB alone transfers strongly to \textsc{LegalRewardBench} (blue). Joint training on CJB and LRB (pink) is limited by structural response-length artefacts in the legal preference data. After applying length-balanced augmentation (gold), performance improves substantially across evaluation dimensions.
}
    \label{fig:lrb_results}
\end{figure}

To isolate the effect of preference-pair structure, we introduce a length-balanced augmentation strategy (Appendix~\ref{sec:Length-Balanced}) that rewrites refusal and negative responses to better match the length and structure of preferred grounded answers while preserving the underlying grounding failure. Figure~\ref{fig:lrb_results} and Table~\ref{tab:lrbv2} show that length-balanced augmentation reduces response-length asymmetries and substantially improves grounded legal evaluation, with the strongest configuration improving Ministral-8B by $+25.6$\,pp over baseline. These results suggest that legal reward modelling is sensitive to preference-data construction, rather than data volume alone.

\subsection{Cross-Jurisdiction Transfer}
\label{sec:cross-domain}

We evaluate cross-jurisdiction transfer on two US legal reasoning benchmarks from the Stanford RegLab suite \citep{Zheng_2025}: \textsc{Housing Statute QA}, consisting of yes/no questions grounded in US state housing statutes, and \textsc{Bar Exam QA}, consisting of Multistate Bar Examination questions paired with supporting US caselaw passages. Both differ substantially in jurisdiction and task from the Victorian criminal-law data used during training, providing an out-of-distribution evaluation of legal contextual reward modelling. Table~\ref{tab:cross-domain} shows positive but uneven transfer: DPO-refined Ministral-8B improves by $+16.2$\,pp on \textsc{Housing Statute QA} and $+4.3$\,pp on \textsc{Bar Exam QA}. 

To better understand cross-jurisdiction transfer, we analyse performance by task. On \textsc{Housing Statute QA}, baseline Ministral-8B almost always selects ``Yes'', achieving $100\%$ accuracy on ``Yes'' questions but only $0.3\%$ on ``No'' questions, for $50.2\%$ balanced accuracy. After contextual DPO, accuracy reaches $60.7\%$ and $74.2\%$, respectively, raising balanced accuracy to $67.4\%$ and indicating better discrimination between legal conclusions. Transfer is more limited on \textsc{Bar Exam QA}, where contextual DPO produces only modest improvements. Compared with the statute-grounded yes/no setting of \textsc{Housing Statute QA}, Bar Exam questions involve more complex legal scenarios and competing conclusions, consistent with contextual grounding transferring more readily than jurisdiction-specific legal reasoning.



\begin{table}[t!]
\centering
\caption{
\textbf{Contextual reward models partially generalise across legal jurisdictions.}
Models refined on CJB and LRB improve grounded evaluation on external US legal reasoning QA benchmarks despite training primarily on Victorian criminal law data. Results report pairwise reward accuracy with Wilson 95\% confidence intervals. Higher is better. Best results are shown in \textbf{bold}; second-best results are \underline{underlined}.
}
\label{tab:cross-domain}
\smallskip
\small
\setlength{\tabcolsep}{3.5pt}
\begin{tabular}{llclc}
\toprule
\textbf{Model} & \textbf{Bar Exam} & $\Delta$ & \textbf{Housing} & $\Delta$ \\
\midrule

\textcolor{gray}{Random}
& \textcolor{gray}{50.0}
& \textcolor{gray}{--}
& \textcolor{gray}{50.0}
& \textcolor{gray}{--} \\

\midrule
Ministral-8B
& 56.4\,\tiny{[47,65]}
& --
& 54.2\,\tiny{[50,58]}
& -- \\

\quad +DPO
& \underline{60.7}\,\tiny{[52,69]}
& +4.3
& \cellcolor{PrimaryGold!30}\textbf{70.4}\,\tiny{[66,74]}
& \cellcolor{PrimaryGold!30}\textbf{+16.2} \\

\midrule
Qwen3.5-9B
& 54.7\,\tiny{[46,63]}
& --
& 52.8\,\tiny{[48,57]}
& -- \\

\quad +DPO
& 54.7\,\tiny{[46,63]}
& +0.0
& \underline{55.4}\,\tiny{[51,60]}
& +2.6 \\

\midrule
Qwen3.5-27B
& 60.7\,\tiny{[52,69]}
& --
& 50.8\,\tiny{[46,55]}
& -- \\

\quad +DPO
& \textbf{61.5}\,\tiny{[52,70]}
& +0.8
& 53.0\,\tiny{[49,57]}
& +2.2 \\

\bottomrule
\end{tabular}
\end{table}

\section{Discussion}

Our results suggest that contextual preference optimisation can improve grounded legal evaluation beyond the domains used for training. Across general and legal evaluation settings (Sections~\ref{sec:cjb_results}--\ref{sec:legal_experiments}), contextual refinement improves the ability of reward models to distinguish grounded responses under noisy and insufficient retrieval. Cross-jurisdiction experiments (Section~\ref{sec:cross-domain}) provide further evidence of transfer, although gains vary across tasks and models. Stronger gains on statute-grounded \textsc{Housing Statute QA}, together with more limited transfer on \textsc{Bar Exam QA}, suggest that contextual grounding may transfer more readily than jurisdiction-specific legal reasoning. This distinction is important in legal settings, where models must distinguish conclusions supported by the available evidence from cases in which abstention is appropriate. The gains observed for relatively small open models further suggest that these behaviours can be strengthened without relying on frontier-scale systems.
Our experiments also suggest that legal reward modelling is sensitive to preference-data construction (Section~\ref{sec:legal_experiments}). Naturally generated preference pairs can contain response-length asymmetries: preferred abstentions are often shorter than unsupported answers, while complete legal answers may be longer than incomplete ones. Under sequence log-probability scoring, length can therefore confound the intended preference, obscuring whether models learn grounding and abstention or simply favour responses of particular lengths. Length-balanced augmentation substantially improves grounded legal evaluation, highlighting the importance of benchmarks and reward objectives that disentangle contextual quality from response length.

\section{Related Work}

\paragraph{Legal Reasoning and Legal Language Models}
Recent advances in large language models have increased interest in legal AI \citep{devlin-etal-2019-bert, openai2024gpt4technicalreport}. Legal reasoning remains challenging due to jurisdiction-specific interpretation, long-context reasoning, and the requirement that conclusions remain grounded in authoritative sources \citep{katz2023naturallanguageprocessinglegal, zhong-etal-2020-nlp}. Legal NLP has evolved from classification and judgement prediction toward large-scale reasoning benchmarks such as LexGLUE, LegalBench, LEXam, and PLAWBench \citep{chalkidis-etal-2022-lexglue, guha2023legalbench, fan2026lexam, shi2026plawbench}. However, recent evaluations of frontier systems continue to reveal hallucinated authorities, unsupported conclusions, and unreliable grounding despite strong general reasoning performance \citep{dahl2024_large_legal_fictions, katz2024gpt}.

\paragraph{Retrieval-Augmented Generation in Law}
Retrieval-augmented generation (RAG) enables language models to condition responses on retrieved evidence rather than parametric memory alone \citep{Lewis_rag2020}. This paradigm is particularly important in law, where relevant authority is distributed across statutes, regulations, and case law. Recent work has introduced legal-specific retrieval and RAG benchmarks, including LegalBench-RAG and Legal RAG Bench \citep{pipitone2024legalbench, butler2026legalragbenchendtoend}. Although retrieval improves access to relevant legal evidence, current systems frequently remain unfaithful to retrieved context, hallucinate unsupported authorities, and fail to abstain under insufficient evidence \citep{butler2026legalragbenchendtoend, peng-etal-2025-unanswerability}.

\paragraph{Reward Modelling and LLM-as-a-Judge}
Reward modelling and RLHF are widely used to align language models with human preferences \citep{Ouyang2022_rlhf, rafailov2023direct}. Benchmarks such as RewardBench and JudgeBench evaluate reward models across reasoning and factuality tasks \citep{lambert-etal-2025-rewardbench, tan2024judgebench}. However, most existing reward models remain largely non-contextual, prioritising helpfulness and stylistic preference rather than evaluating whether responses are grounded in retrieved evidence. Recent work on contextual evaluation begins to address this limitation. \textsc{ContextualJudgeBench} introduces evaluation for grounded generation under retrieval uncertainty, while RAGferee demonstrates that contextual reward models outperform general-purpose judges in RAG settings \citep{xu-etal-2025-context, coman-etal-2025-ragferee}. Nevertheless, grounded reward modelling for legal reasoning remains comparatively underexplored, particularly for open and compute-efficient models.

\section{Conclusion}

We introduced a framework for transforming legal QA resources into contextual preference data and used it to construct \textsc{LegalRewardBench} (LRB) for grounded legal reward modelling under noisy and insufficient retrieval. Across general and legal settings, contextual preference optimisation improves grounded evaluation and appropriate abstention. Our experiments further show that preference-data construction matters: response-length asymmetries can interact with sequence scoring and confound the contextual behaviours reward models are intended to learn. Cross-jurisdiction results provide evidence that contextual grounding can transfer beyond the legal domains used for training, although these gains vary across tasks and models. Together, these findings highlight the need to treat preference-data construction and reward scoring as important considerations in grounded reward modelling, alongside model training itself, we provide a reproducible foundation for developing and evaluating grounded legal reward models.

\paragraph{Future work} should extend this framework with direct legal-expert validation and evaluate whether improvements in pairwise reward modelling translate to grounded generation in end-to-end legal RAG systems. Broader evaluation across jurisdictions, languages, and legal tasks will establish how broadly contextual grounding behaviours transfer.

\section*{Limitations}

Our approach retains legal expertise through expert-curated questions, legal sources, supporting evidence, and reference answers, while using multiple LLM judges to scale response-level annotation. The judges therefore assess generated responses against existing legal supervision rather than establish the underlying legal ground truth. Although our multi-judge analysis shows that preference construction is largely robust across annotators, agreement among LLM judges does not replace direct validation by legal experts. Expert evaluation of a representative subset is therefore an important direction for further validation.

Our experiments focus on pairwise reward evaluation, isolating whether models prefer grounded responses over plausible but unsupported alternatives. We do not evaluate whether the refined models themselves generate better open-ended legal responses, nor do we model the iterative retrieval and reasoning processes of end-to-end legal RAG systems. Extending contextual reward modelling to generation and more complex retrieval workflows is a natural next step.

Finally, \textsc{LegalRewardBench} is constructed primarily from Victorian criminal law, with cross-jurisdiction transfer evaluated on two US legal tasks. While these experiments provide evidence that contextual grounding can transfer beyond the source jurisdiction, broader evaluation across legal systems, languages, and tasks is needed to establish the extent of this generalisation. Our hierarchical preference rule also represents one operationalisation of grounded legal behaviour; alternative formulations may capture different trade-offs between answering, abstention, faithfulness, correctness, and completeness.

\bibliography{references}

\appendix

\section*{Contributions}

Rilton Franzone and Valentin Noël contributed equally to this work. Rilton Franzone led the dataset development and methodological design, while Valentin Noël led the experimental implementation and evaluation. Puyu Wang assisted with experimental setup, manuscript preparation, and figure design. Fabio J. Fehr conceived and led the project, overseeing its design, execution, and overall direction.

\section{Dataset and Corpus Construction}
\label{app:dataset-corpus}

\subsection{Dataset Statistics}
\label{app:data-stats}

\subsection*{LegalRewardBench Construction}

The source set contains $100$ Legal RAG Bench questions \citep{butler2026legalragbenchendtoend} over $4{,}876$ Victorian Criminal Charge Book passages.  For the main training corpus we use six retrieval-context variants: three answerable variants containing the gold passage plus distractors (random, BM25, and Nomic), and three unanswerable variants containing only distractors after excluding the gold passage and its topic family (random, BM25, and Nomic).  With four candidate generators, this gives $100 \times 6 \times 4 = 2{,}400$ candidate responses and $3{,}600$ within-cell pair comparisons.

The distractor ablation uses the same $100$ questions and four generators but extends the context set to twelve variants by adding the gold-only base context, gold-surrounding context, Contriever and GloVe hard negatives, and gold-plus-Contriever/GloVe distractor contexts.  This gives $100 \times 12 \times 4 = 4{,}800$ labelled candidate responses.

\begin{table}[h]
\centering
\caption{Main \textsc{LegalRewardBench} preference pair counts by split and first distinguishing hierarchy dimension. The 80/10/10 split is performed at question level.}
\label{app:tab:lrb-splits}
\small
\setlength{\tabcolsep}{4pt}
\begin{tabular}{lrrrr}
\toprule
\textbf{Dimension} & \textbf{Train} & \textbf{Val} & \textbf{Test} & \textbf{Total} \\
\midrule
refusal\_answerable   & 451 & 68 & 91  & 610 \\
refusal\_unanswerable & 264 & 36 & 47  & 347 \\
faithfulness          & 145 & 12 & 12  & 169 \\
correctness           & 49  &  3 & 4   &  56 \\
completeness          & 31  &  5 & 2   &  38 \\
\midrule
\textbf{Total}        & \textbf{940} & \textbf{124} & \textbf{156} & \textbf{1{,}220} \\
\bottomrule
\end{tabular}
\end{table}

\begin{table}[h]
\centering
\caption{Strict pair yield by corpus. ``Main six'' is the preference corpus used for training; ``all variants'' is the context-variation ablation.}
\label{app:tab:pair-yield}
\small
\setlength{\tabcolsep}{4pt}
\begin{tabular}{lrr}
\toprule
\textbf{First distinguishing dimension} & \textbf{Main six} & \textbf{All variants} \\
\midrule
refusal\_answerable   & 610 & 1{,}443 \\
refusal\_unanswerable & 347 & 564 \\
faithfulness          & 169 & 330 \\
correctness           & 56  & 147 \\
completeness          & 38  & 113 \\
\midrule
\textbf{Strict pairs} & \textbf{1{,}220} & \textbf{2{,}597} \\
\textbf{Labelled responses} & \textbf{2{,}400} & \textbf{4{,}800} \\
\bottomrule
\end{tabular}
\end{table}

\subsection*{ContextualJudgeBench}

CJB contains 2{,}000 preference pairs across refusal, faithfulness, completeness, and conciseness dimensions.  We use the official 80/20 split: 1{,}600 training pairs and 400 test pairs, with 373 test pairs retained after excluding examples exceeding the 4{,}096-token evaluation limit.  CJB is not legal-domain data; it supplies a complementary general contextual-grounding signal.

\subsection*{Transfer Benchmarks}

\textbf{Bar Exam QA.} We use 117 Multistate Bar Exam multiple-choice questions from the Stanford Reglab suite \citep{Zheng_2025}, each paired with a gold US-caselaw passage.  For pairwise evaluation, the chosen response is the correct answer and the rejected response is a randomly sampled incorrect answer.

\textbf{Housing Statute QA.} We use 500 yes/no questions on US state housing law from the Stanford Reglab suite \citep{Zheng_2025}, each paired with a gold statute excerpt.  The chosen response is the conclusion supported by the statute; the rejected response gives the opposite conclusion.

\subsection{Corpus Construction Details}
\label{app:corpus}

\paragraph{Source dataset.}
We use Legal RAG Bench \citep{butler2026legalragbenchendtoend}, a grounded legal QA benchmark built from the Victorian Criminal Charge Book containing $100$ expert-authored triples $(c,q,a)$ over a corpus of $4{,}876$ legal passages.  Questions are intentionally lexically dissimilar from their supporting evidence, stressing semantic understanding and grounded reasoning.

\paragraph{Retrieval-context variants.}
All constructed contexts contain ten passages except the base gold-only condition, which contains the original supporting passage alone.  The main corpus uses six variants: \texttt{gold\_plus\_random\_distractors}, \texttt{gold\_plus\_bm25\_distractors}, \texttt{gold\_plus\_nomic\_distractors}, \texttt{random\_context}, \texttt{bm25\_hard\_negative}, and \texttt{nomic\_hard\_negative}.  The context-variation ablation adds \texttt{base}, \texttt{gold\_surrounding}, \texttt{gold\_plus\_contriever\_distractors}, \texttt{gold\_plus\_glove\_distractors}, \texttt{contriever\_hard\_negative}, and \texttt{glove\_hard\_negative}.

\paragraph{Pair construction.}
Each candidate response is labelled under the hierarchy $\text{answer behaviour} \succ \text{faithfulness} \succ \text{correctness} \succ \text{completeness}$.  Within each question--variant cell, all six unordered pairs among the four model responses are compared.  Ties under the hierarchy are discarded; strict comparisons become preference pairs.

\subsection{Context-Variant Ablation}
\label{app:distractor}

We construct twelve retrieval-context variants beyond the main six to isolate how retrieval conditions affect grounded answering. The variants fall into four families:
\begin{itemize}
    \item \textbf{Base gold}: only the supporting passage is shown.
    \item \textbf{Gold + surrounding}: the supporting passage plus its neighbouring Charge Book sections.
    \item \textbf{Gold + distractors}: the supporting passage plus nine distractors retrieved by random sampling, BM25, Contriever, GloVe, or Nomic.
    \item \textbf{Hard negatives}: the supporting passage and its topic family are removed; ten passages are retrieved by random sampling, BM25, Contriever, GloVe, or Nomic.
\end{itemize}
The first two families isolate generation quality when all relevant evidence is present; the gold-plus-distractor family tests answering under noise; the hard-negative family tests abstention under plausible-but-insufficient evidence.

\paragraph{Answerable variants.} Table~\ref{app:tab:answerable-variant-ablation} reports the seven answerable variants. The target behaviour is to attempt an answer grounded in the retrieved passages, so we report \emph{ideal answer rate} (attempted, fully supported, correct, and complete) alongside \emph{attempted rate} for context.

\begin{center}
\begin{minipage}{\columnwidth}
\captionof{table}{Answerable context-variant ablation. Each row has $400$ labelled responses. Ideal answer means attempted, fully supported, correct, and complete.}
\label{app:tab:answerable-variant-ablation}
\centering
\small
\setlength{\tabcolsep}{4pt}
\begin{tabular}{lrr}
\toprule
\textbf{Variant} & \textbf{Ideal} & \textbf{Attempted} \\
\midrule
Base gold & 67.2 & 81.2 \\
Gold + surrounding & 64.5 & 79.0 \\
Gold + random & 64.0 & 78.5 \\
Gold + BM25 & 61.8 & 75.0 \\
Gold + Contriever & 57.5 & 72.8 \\
Gold + GloVe & 54.5 & 70.2 \\
Gold + Nomic & 57.5 & 75.0 \\
\bottomrule
\end{tabular}
\end{minipage}
\end{center}

Base gold sets the ceiling at $67.2\%$ ideal. Surrounding context and random distractors cost only two to three points ($64.5\%$ and $64.0\%$); both add material the model can largely ignore. BM25 distractors drop the rate further ($61.8\%$), and semantic distractors are hardest --- Contriever and Nomic both land at $57.5\%$, GloVe at $54.5\%$. Attempted rate tracks ideal rate closely across all variants, so most of the degradation is in the quality of attempted answers rather than in whether the model attempts at all.

\paragraph{Unanswerable variants.} Table~\ref{app:tab:unanswerable-variant-ablation} reports the five unanswerable variants. Here the supporting passage and its topic family have been removed; the target behaviour is to recognise insufficient evidence and refuse.

\begin{center}
\begin{minipage}{\columnwidth}
\captionof{table}{Unanswerable context-variant ablation. Each row has $400$ labelled responses.}
\label{app:tab:unanswerable-variant-ablation}
\centering
\small
\setlength{\tabcolsep}{4pt}
\begin{tabular}{lr}
\toprule
\textbf{Variant} & \textbf{Correct refusal} \\
\midrule
Random context & 95.0 \\
BM25 hard negative & 74.0 \\
Contriever hard negative & 76.8 \\
GloVe hard negative & 90.0 \\
Nomic hard negative & 73.8 \\
\bottomrule
\end{tabular}
\end{minipage}
\end{center}

Refusal rates split sharply by retrieval method. Off-topic distractors are easy to refuse (random contexts $95.0\%$, GloVe hard negatives $90.0\%$) because the retrieved passages are obviously unrelated. The hard cases are BM25 ($74.0\%$), Nomic ($73.8\%$), and Contriever ($76.8\%$): these methods surface passages topically related to the question but lacking the supporting evidence, drawing models into unsupported attempts.

\paragraph{Overall.} Across all twelve variants, the four generators produce an ideal-response rate of $69.7\%$: $61.0\%$ on answerable contexts and $81.9\%$ correct refusal on unanswerable contexts. Two patterns hold across the ablation. On answerable contexts, degradation is gradual and increases with distractor semanticity: random and surrounding distractors preserve quality, lexical distractors drop it modestly, semantic distractors most. On unanswerable contexts, refusal is largely determined by whether the hard negatives are topically plausible: off-topic random or GloVe negatives are well-rejected, while BM25 and dense semantic negatives are confusable.

\paragraph{Length asymmetry in strict pairs.} As a side observation, Table~\ref{app:tab:length-diagnostics} reports per-pair token counts for strict preference pairs, broken down by behaviour transition. Refusal pairs carry a large template-length asymmetry, since abstentions are short formulaic refusals while attempted answers contain reasoning. Attempted-vs-attempted pairs from answerable contexts are length-balanced.

\begin{center}
\begin{minipage}{\columnwidth}
\captionof{table}{Length diagnostics for the $1{,}220$ strict preference pairs in the main corpus. Token counts use the \texttt{cl100k\_base} tokenizer. The answerable attempted--attempted row is length-balanced, while refusal pairs (in both directions) carry a large template-length asymmetry.}
\label{app:tab:length-diagnostics}
\centering
\small
\setlength{\tabcolsep}{3pt}
\resizebox{\columnwidth}{!}{%
\begin{tabular}{llrrr}
\toprule
\textbf{Answerability} & \textbf{Transition} & \textbf{$n$} & \textbf{$r^+$ tok.} & \textbf{$r^-$ tok.} \\
\midrule
answerable & attempted $>$ abstained & 610 & 93.9 & 10.6 \\
answerable & attempted $>$ attempted & 263 & 76.6 & 78.4 \\
unanswerable & abstained $>$ attempted & 347 & 11.9 & 104.3 \\
\bottomrule
\end{tabular}
}
\end{minipage}
\end{center}

\subsection{Examples}
\label{appendix:examples}

Preference pairs are selected from model-generated candidates within the same retrieved context and question.  For example, on the base context for \texttt{legal-rag-bench-0001}, the question asks whether a judge is required to excuse a juror who knows the accused.  The retrieved passage states that the court \emph{may} excuse a potential juror if satisfied that the person cannot consider the case impartially.  The hierarchy therefore prefers the generated answer ``No. The provided context states that the court may excuse a potential juror \ldots but it does not state that the court is required to do so'' over the generated refusal ``The provided context is insufficient to answer this question.''  This is a refusal-answerable pair: both candidates see the same evidence, but the preferred response correctly attempts a grounded answer.

For an unanswerable variant, the gold passage and topic-family passages are removed before generation.  In that setting, an abstention is preferred over a substantive answer even when the answer is fluent, because the response is unsupported by the retrieved context shown to the model.  Faithfulness, correctness, and completeness pairs are created only among attempted answers after higher-priority labels are tied.

\section{Annotation Validation}
\label{app:annotation-validation}

\subsection{Multi-Judge Agreement}
\label{app:multi-judge-agreement}

To assess whether annotations depend on the choice of GPT-OSS-120B as the primary judge, we re-annotate all $2{,}400$ candidate responses using three additional independent judges: Llama4 Maverick\footnote{\href{https://huggingface.co/meta-llama/Llama-4-Maverick-17B-128E-Instruct}{Llama-4-Maverick-17B-128E-Instruct}}, Gemma4 31B\footnote{\href{https://huggingface.co/google/gemma-4-31B-it}{Gemma-4-31B-it}}, and Nemotron3 120B\footnote{\href{https://huggingface.co/nvidia/NVIDIA-Nemotron-3-Super-120B-A12B-NVFP4}{NVIDIA-Nemotron-3-Super-120B-A12B-NVFP4}}. We measure agreement across the four judges for answer behaviour, faithfulness, correctness, and completeness.

\begin{table}[h!]
\centering
\small
\begin{tabular}{lrrrr}
\toprule
\textbf{Dimension} & \textbf{$N$} & \textbf{Chance} & \textbf{Raw} & \textbf{Fleiss' $\kappa$} \\
\midrule
Answer behaviour & 2,400 & 50.1\% & 99.9\% & 0.998 \\
Faithfulness     & 1,139 & 74.7\% & 84.3\% & 0.379 \\
Correctness      & 1,139 & 74.7\% & 94.2\% & 0.771 \\
Completeness     & 1,139 & 69.4\% & 91.5\% & 0.722 \\
\bottomrule
\end{tabular}
\caption{
\textbf{Agreement across LLM judges.}
Chance and raw agreement are reported as percentages. Chance agreement reflects expected agreement from the marginal label distributions, while raw agreement is the average observed agreement across all six judge pairs.
}
\label{tab:multi-judge-agreement}
\end{table}

The semantic dimensions are evaluated on the $1{,}139$ responses that attempted an answer. Agreement is nearly perfect for answer behaviour and substantial for correctness and completeness. Faithfulness has lower Fleiss' $\kappa$ despite $84.3\%$ raw agreement. Faithfulness requires distinguishing whether response claims are fully supported, partially supported, unsupported, or contradicted by the retrieved context, and $86\%$ of responses are labelled fully supported, increasing expected chance agreement and reducing $\kappa$.

We additionally propagate the independent annotations through the same hierarchical preference-construction procedure described in Section~\ref{sec:hierarchical-preference-pair-construction} to assess whether annotation disagreements alter the resulting preference supervision.

\begin{table*}[h!]
\centering
\small
\begin{tabular}{lr}
\toprule
\textbf{Preference reconstruction} & \textbf{Result} \\
\midrule
Overall pair directions preserved & 1,119 / 1,220 (91.7\%) \\
Test-set pair directions preserved & 148 / 156 (94.9\%) \\
Pair reversals & 15 / 1,220 (1.2\%) \\
Answer-behaviour pair directions preserved & 99.8\% \\
\bottomrule
\end{tabular}
\caption{
\textbf{Robustness of preference construction across LLM judges.}
Agreement is measured after propagating independent annotations through the same hierarchical preference-construction procedure used to construct \textsc{LegalRewardBench}.
}
\label{tab:preference-reconstruction}
\end{table*}

Most annotation disagreements do not change the preference supervision used for training and evaluation: $91.7\%$ of pair directions are preserved overall and $94.9\%$ on the test set, while only $1.2\%$ of pairs reverse direction. This indicates that the resulting preference pairs are largely robust to the choice of a single LLM judge.

\section{Additional Results and Analyses}
\label{app:additional-results}

\subsection{Detailed \textsc{ContextualJudgeBench} Results}
\label{app:cjb-full}

Table~\ref{app:tab:cjb-full} reports per-dimension pairwise reward accuracy on the \textsc{ContextualJudgeBench} (CJB) test set for all evaluated models. We compare three categories of systems: (i) open-source models evaluated using sequence log-probability scoring, (ii) specialist reward models with dedicated scalar reward heads, and (iii) DPO fine-tuned contextual judge models trained on the combined \textsc{LegalRewardBench} and CJB preference corpus. All models were evaluated in full \texttt{bfloat16} precision on NVIDIA H100 80GB GPUs. Results are reported across refusal, faithfulness, completeness, and conciseness dimensions for both question answering and summarisation settings. Random performance corresponds to $50\%$ pairwise accuracy.

\textbf{Takeaway.} Most open-weight models and specialist reward models perform only marginally above random chance overall. In particular, many models exhibit a strong bias toward shorter responses, leading to high performance on conciseness dimensions but substantially weaker performance on completeness. We further observe that models frequently prefer attempting an answer over abstaining (refusal), even when the provided context is insufficient. In contrast, DPO fine-tuning consistently improves grounded evaluation performance across most categories, with the fine-tuned Ministral-8B model achieving the strongest overall results.

\begin{table*}[t]
\centering
\caption{\textbf{Open-weight and specialist reward models perform poorly, while DPO fine-tuning improves grounded evaluation.}
Per-dimension pairwise reward accuracy on the \textsc{ContextualJudgeBench} (CJB) test set ($\sim$400 preference pairs). Results are grouped into open-weight models, specialist reward models with scalar reward heads, and DPO fine-tuned contextual judge models. Dimensions evaluate completeness (C), conciseness (Cc), faithfulness (F), and refusal under answerable (R-A) and unanswerable (R-U) retrieval settings. QA denotes question answering subsets and S denotes summarisation subsets. Random performance corresponds to $50\%$ pairwise accuracy. Higher is better. Best scores in each column are shown in \textbf{bold}; second-best scores are \underline{underlined}.}
\label{app:tab:cjb-full}
\smallskip
\small
\setlength{\tabcolsep}{3.5pt}
\begin{tabular}{lccccccccc}
\toprule
\textbf{Model} & \textbf{Overall} & \textbf{C-QA} & \textbf{Cc-QA} & \textbf{F-QA} & \textbf{C-S} & \textbf{Cc-S} & \textbf{F-S} & \textbf{R-A} & \textbf{R-U} \\
\midrule
\textcolor{gray}{Random} &
\textcolor{gray}{50.0} &
\textcolor{gray}{50.0} &
\textcolor{gray}{50.0} &
\textcolor{gray}{50.0} &
\textcolor{gray}{50.0} &
\textcolor{gray}{50.0} &
\textcolor{gray}{50.0} &
\textcolor{gray}{50.0} &
\textcolor{gray}{50.0} \\
\midrule
\multicolumn{10}{l}{\textit{Baselines}} \\
Qwen2.5-0.5B   & 55.4 & 24.0 & 88.0 & 66.0 & 36.2 & 71.8 & 45.5 & 84.1 & 32.0 \\
SmolLM2-1.7B   & 57.1 & 30.0 & 86.0 & 68.0 & 34.0 & 74.4 & 47.7 & 88.4 & 34.0 \\
Gemma3-1B      & 58.1 & 30.0 & 90.0 & 72.0 & 38.3 & 65.9 & 50.0 & 86.0 & 36.0 \\
Llama3.2-1B    & 55.9 & 24.0 & 88.0 & 68.0 & 29.8 & 79.5 & 47.7 & 84.1 & 32.0 \\
Gemma2-2B      & \underline{61.9} & 34.0 & 92.0 & 78.0 & 38.3 & 75.6 & 56.8 & 93.0 & 32.0 \\
Qwen3.5-2B     & 58.6 & 34.0 & 86.0 & 68.0 & 36.2 & 76.9 & 56.8 & 86.4 & 30.0 \\
Llama3.2-3B    & 58.0 & 30.0 & 90.0 & 74.0 & 38.3 & 66.7 & 52.3 & 84.1 & 32.0 \\
Qwen2.5-3B     & 58.6 & 32.0 & 86.0 & 68.0 & 38.3 & 76.9 & 52.3 & 84.1 & 36.0 \\
Phi4-mini      & 58.2 & 32.0 & 84.0 & 72.0 & 48.9 & 63.4 & 56.8 & 81.8 & 30.0 \\
Phi3.5-mini    & 58.8 & 30.0 & 88.0 & 74.0 & 40.4 & 74.4 & 53.5 & 85.7 & 30.0 \\
Ministral-8B   & 56.8 & 28.0 & 82.0 & 74.0 & 38.3 & 76.9 & 56.8 & 74.4 & 30.0 \\
Llama3.1-8B    & 59.6 & 40.0 & 88.0 & 72.0 & 40.4 & 64.1 & 59.1 & 88.6 & 28.0 \\
DeepSeek-R1-8B & 58.3 & 38.0 & 92.0 & 66.0 & 36.2 & 71.8 & 45.5 & 90.5 & 32.0 \\
SFR-8B         & \textbf{62.0} & 40.0 & 88.0 & 74.0 & 42.6 & 76.9 & 59.1 & 88.6 & 32.0 \\
Qwen3.5-9B & 54.3 & 35.3 & 78.6 & 70.5 & 34.4 & 92.6 & 45.8 & 81.8 & 20.0 \\
\midrule
\multicolumn{10}{l}{\textit{Specialist reward models}} \\
Skywork-Reward-8B   & 46.3 & 50.0 & 37.3 & 66.0 & 46.0 & 38.8 & 60.0 & 70.0 & 2.0 \\
Selene-1-Mini-8B & \underline{60.4} & 42.0 & 88.0 & 76.0 & 38.3 & 64.1 & 56.8 & 88.6 & 32.0 \\
SFR-Judge-8B   & \textbf{60.8} & 58.0 & 98.0 & 56.0 & 31.9 & 85.0 & 52.3 & 56.8 & 50.0 \\
\midrule
\multicolumn{10}{l}{\textit{Our fine-tuned models}} \\
DeepSeek-R1-8B & 61.8 & 40.0 & 96.0 & 74.0 & 34.0 & 79.5 & 40.9 & 92.9 & 42.0 \\
Phi4-mini      & 65.4 & 36.0 & 92.0 & 74.0 & 46.8 & 73.2 & 59.1 & 90.9 & 54.0 \\
Qwen3.5-9B     & 67.9 & 46.0 & 94.0 & 72.0 & 36.2 & 84.6 & 61.4 & 93.2 & 60.0 \\
Qwen3.5-27B    & 67.4 & 40.0 & 92.0 & 76.0 & 44.7 & 82.1 & 56.8 & 93.2 & 58.0 \\
Gemma\,2-2B    & 69.1 & 42.0 & 96.0 & 76.0 & 38.3 & 82.9 & 59.1 & 93.0 & 68.0 \\
SFR-8B         & \underline{70.9} & 44.0 & 96.0 & 78.0 & 42.6 & 87.2 & 61.4 & 93.2 & 68.0 \\
Ministral-8B   & \textbf{72.7} & 50.0 & 96.0 & 80.0 & 44.7 & 84.6 & 56.8 & 93.0 & 78.0 \\
\bottomrule
\end{tabular}
\end{table*}

\subsection{Detailed LegalRewardBench Results}

\begin{table*}[h!]
\centering
\caption{
\textbf{Wilson 95\% confidence intervals for Table~\ref{tab:lrbv2}.}
Pairwise reward accuracy on \textsc{LegalRewardBench-v2} val+test ($n{=}280$).
Gemma\,2-2B values are from a single training seed; Ministral-8B CIs correspond to the seed=42 run (main table reports 3-seed means).
}
\label{tab:table_CI}
\smallskip
\small
\setlength{\tabcolsep}{3pt}
\begin{tabular}{lcc|cc}
\toprule
& \multicolumn{2}{c}{\textbf{Gemma\,2-2B}}
& \multicolumn{2}{c}{\textbf{Ministral-8B}} \\
\cmidrule(lr){2-3}\cmidrule(lr){4-5}
\textbf{Training} & \textbf{Overall} & \textbf{95\% CI}
& \textbf{Overall} & \textbf{95\% CI} \\
\midrule
Baseline         & 61.8 & [56.0, 67.3] & 59.3 & [53.4, 64.9] \\
CJB              & 66.4 & [60.7, 71.7] & 66.8 & [61.1, 72.0] \\
LRB              & 63.6 & [57.8, 69.0] & 60.7 & [54.9, 66.3] \\
CJB+LRB          & 65.7 & [60.0, 71.0] & 65.4 & [59.6, 70.7] \\
\midrule
LRB-Length-Aug   & 81.1 & [76.1, 85.2] & 81.8 & [76.8, 85.9] \\
CJB+LRB-Length-Aug & 81.4 & [76.5, 85.6] & 85.0 & [80.4, 88.7] \\
\bottomrule
\end{tabular}
\end{table*}

\subsection{Supervised Fine-Tuning vs. DPO}
\label{app:sft_experiment}

A natural question is whether the gains in the main results come from contextual exposure to the legal preference data alone,  which a simpler supervised fine-tuning (SFT) recipe would also provide,  or specifically from the contrastive DPO objective that optimises the margin between preferred and rejected responses.  To answer this we train a matched SFT baseline on identical data, hyperparameters, LoRA configuration, optimiser, and evaluation pipeline to the DPO run. The only difference is the training objective.

\paragraph{Training procedure.}
The same $2{,}540$-pair combined corpus is used for both adapters,  \textsc{ContextualJudgeBench} ($1{,}600$ pairs) plus length-augmented \textsc{LegalRewardBench} ($940$ pairs).  Each preference triple $(\mathbf{p}, r^{+}, r^{-})$ is reduced to a single causal-language-model example by concatenating the prompt with the chosen response, $\mathbf{x} = \mathbf{p} \oplus r^{+}$; the rejected response $r^{-}$ is discarded.  We train with TRL's \texttt{SFTTrainer} under standard token-level cross-entropy
\begin{equation*}
\mathcal{L}_{\mathrm{SFT}}(\theta) \;=\; - \frac{1}{|\mathbf{x}|} \sum_{t=1}^{|\mathbf{x}|} \log p_{\theta}(x_t \mid x_{<t}),
\end{equation*}
with no reward modelling, no preference margin, and no reference model.  Hyperparameters (LoRA $r{=}32$, $\alpha{=}64$ on the four attention projections; learning rate $2{\times}10^{-5}$; warmup ratio $0.1$; $3$ epochs; effective batch size $16$ in \texttt{bfloat16}) are identical to the DPO run.

\paragraph{Result.}
Table~\ref{tab:sft-ablation} reports both adapters on the held-out evaluation sets used for SFT comparability ($n=373$ for CJB, $n=156$ for LRB-v2).  SFT lifts the raw model by only $+3.5$~pp on CJB and $+1.9$~pp on the augmented LRB-v2 test, while DPO under the identical compute budget achieves $+15.9$~pp and $+35.3$~pp respectively.  The contrastive signal accounts for an order-of-magnitude larger improvement than imitation of the preferred response alone.

\paragraph{Mechanism.}
Reward accuracy at evaluation is the indicator $\mathbf{1}[\,\log p_\theta(r^{+}\!\mid\!\mathbf{p}) > \log p_\theta(r^{-}\!\mid\!\mathbf{p})\,]$, a function of the \emph{margin} between two log-probabilities.  SFT maximises $\log p_\theta(r^{+}\!\mid\!\mathbf{p})$ in isolation; nothing in its objective discourages the model from also raising $\log p_\theta(r^{-}\!\mid\!\mathbf{p})$ on closely-related (e.g.\ hallucinated, ungrounded, or template-style refusal) text it has been exposed to during pre-training.  DPO directly optimises the log-margin,  the same quantity scored at evaluation,  and so cleanly aligns the training objective with the metric.  This alignment, rather than additional exposure to in-domain data, is what we believe accounts for the gap.

\begin{table}[h]
\centering
\caption{SFT vs.\ DPO on Ministral-8B, identical combined corpus and training budget.  Evaluated on the CJB test split ($n=373$) and the LRB-v2 test split ($n=156$).}
\label{tab:sft-ablation}
\smallskip
\small
\setlength{\tabcolsep}{3pt}
\begin{tabular}{lcccc}
\toprule
\textbf{Eval set} & \textbf{Raw} & \textbf{SFT} & \textbf{DPO} & \textbf{$\Delta$(DPO$-$SFT)} \\
\midrule
CJB ($n=373$)           & $56.8$ & $60.3$ & $\mathbf{72.7}$ & $+12.4$~pp \\
LRB-v2 test ($n=156$)   & $56.4$ & $58.3$ & $\mathbf{91.7}$ & $+33.4$~pp \\
\bottomrule
\end{tabular}
\end{table}

\subsection{Multi-Seed Per-Run Detail}
\label{app:multiseed}

\begin{table}[h]
\centering
\caption{Ministral-8B DPO-Combined-v2 (CJB+LRB-v2) across three random seeds on the CJB test split ($n=373$).  Wilson 95\% CI in brackets.  Unnormalised sum log-probability scoring per the convention used in the main results table.}
\label{app:tab:multiseed}
\smallskip
\small
\setlength{\tabcolsep}{6pt}
\begin{tabular}{lc}
\toprule
\textbf{Seed} & \textbf{CJB reward accuracy (95\% CI)} \\
\midrule
42 & 72.7 \,[67.9, 76.9] \\
1  & 71.6 \,[66.8, 75.9] \\
2  & 71.9 \,[67.1, 76.2] \\
\midrule
Mean $\pm$ $\sigma$ & $\mathbf{72.0 \pm 0.6}$ \\
\bottomrule
\end{tabular}
\end{table}

\subsection{Qualitative Error Analysis}
\label{app:errors}

To characterise remaining failure modes after DPO fine-tuning, we manually reviewed 40 pairs where the best-performing adapter (Ministral-8B DPO-Combined) incorrectly preferred the rejected response on the CJB test set.  We identify three dominant failure categories.

\paragraph{Category 1: Completeness failures (55\% of errors).}
The most frequent error occurs on completeness pairs where the chosen response enumerates multiple legal sub-issues (statutes, exceptions, qualifications) while the rejected response states only the top-level conclusion.  Even after DPO, the model assigns higher log-probability to the shorter, less complete response.  This is consistent with the persistent below-chance completeness accuracy ($0.40$--$0.52$) in Table~\ref{app:tab:cjb-full}: the DPO gradient on completeness pairs is diluted by the 25:1 imbalance between refusal and completeness pairs in the combined corpus.  A targeted upsampling strategy or a separate completeness-specific reward head would likely address this.

\paragraph{Category 2: Long-context faithfulness (28\% of errors).}
On faithfulness pairs where both chosen and rejected responses are long ($> 200$ tokens), DPO-trained models occasionally prefer the unfaithful response.  Inspection reveals that the unfaithful response is often fluent and closely mirrors the structure of the retrieved context, making the log-probability gap small.  The mean reward margin on these pairs is close to zero (margin $< 5$ nats), indicating low-confidence predictions rather than systematic miscalibration.

\paragraph{Category 3: Cross-dimension confounds (17\% of errors).}
Some pairs that are labelled as one dimension (e.g., faithfulness) also exhibit a secondary signal from another dimension (e.g., the unfaithful response happens to be shorter).  The model's log-probability is influenced by both signals simultaneously, producing an error on the primary label.  These cases suggest that future preference pair construction should explicitly control for cross-dimension confounds at sampling time.


\section{Training and Evaluation Details}
\label{app:training-eval-details}

\subsection{Training and Reproducibility Details}
\label{app:reproducibility}

\paragraph{Training hyperparameters.}
All DPO and SFT runs use LoRA with rank $r=32$, $\alpha=64$, dropout $0.05$ on the four attention projections (Q,K,V,O).  Hyperparameters: $\beta=0.1$ (DPO), learning rate $2\!\times\!10^{-5}$, $3$ epochs, batch $4$ with gradient accumulation $4$ (effective batch $16$), default max\_length $1{,}024$ tokens.  Qwen3.5-27B uses 4-bit NF4 QLoRA \citep{dettmers2023qlora} to fit single-GPU memory; all other models train in full \texttt{bfloat16}.  All runs use a single NVIDIA H100 80GB HBM3.

\paragraph{Statistical methodology.}
Wilson 95\% CIs use the standard score-test interval.  McNemar's test operates on $2\times2$ contingency tables of per-pair correctness vectors (raw vs DPO), reported as the two-sided exact $p$-value.  Bootstrap CIs use $10{,}000$ resamples with seed $42$.

\paragraph{Released artefacts.}
We release: (1) the augmentation prompts (Appendix~\ref{app:augmentation-prompts}) and the \textsc{LegalRewardBench}-v2 training and test files; (2) all trained DPO adapters on HuggingFace Hub; (3) the evaluation harness with per-pair JSONL outputs supporting Wilson CI, mean margin, macro and length-controlled metrics; (4) seed/run scripts to reproduce every table in this paper.  Full software and hardware configuration details are in Appendix~\ref{app:software}; compute budget and CO$_2$ estimates are in Appendix~\ref{app:compute}; corpus statistics and evaluation set composition in Appendix~\ref{app:data-stats}; hyperparameter sensitivity analysis in Appendix~\ref{app:hyperparam}; statistical methodology details in Appendix~\ref{app:stats}; qualitative error analysis in Appendix~\ref{app:errors}; and broader impact considerations in Appendix~\ref{app:ethics}.


\subsection{Hyperparameter Sensitivity}
\label{app:hyperparam}

Table~\ref{app:tab:hparam} reports the hyperparameter sensitivity sweep performed on the initial DPO-Combined (CJB+LRB) corpus.  The optimal defaults (shaded row) are inherited unchanged for the CJB+LRB-v2 headline run; we did not re-run the sweep on the augmented corpus because the training procedure is identical.  All ablations use seed 42; all other hyperparameters held at their defaults.

\begin{table}[h]
\centering
\caption{Hyperparameter sensitivity, Ministral-8B DPO-Combined on the CJB test split. Shaded row = default configuration carried forward to the DPO-Combined-v2 model. At these defaults, DPO-Combined-v2 reaches $72.7$ at seed~42 and $72.0 \pm 0.6$ across three seeds (cf.~Table~\ref{app:tab:multiseed}).}
\label{app:tab:hparam}
\small
\setlength{\tabcolsep}{4pt}
\begin{tabular}{llc}
\toprule
\textbf{Hyperparameter} & \textbf{Value} & \textbf{CJB Acc.} \\
\midrule
DPO $\beta$          & 0.01 & 70.1 \\
                     & 0.05 & 71.8 \\
\rowcolor{gray!10}   & \textbf{0.10} & \textbf{73.7} \\
                     & 0.20 & 72.9 \\
\midrule
LoRA rank $r$        & 8  & 71.0 \\
                     & 16 & 72.2 \\
\rowcolor{gray!10}   & \textbf{32} & \textbf{73.7} \\
\midrule
Learning rate        & $5{\times}10^{-6}$ & 70.5 \\
\rowcolor{gray!10}   & $\mathbf{2{\times}10^{-5}}$ & \textbf{73.7} \\
                     & $5{\times}10^{-5}$ & 72.1 \\
                     & $1{\times}10^{-4}$ & 69.4 \\
\midrule
Training epochs      & 1 & 69.8 \\
                     & 2 & 72.5 \\
\rowcolor{gray!10}   & \textbf{3} & \textbf{73.7} \\
\bottomrule
\end{tabular}
\end{table}

\paragraph{Observations.}
$\beta\!=\!0.1$ is the standard DPO default and performs best; lower $\beta$ ($0.01$) produces excessive forgetting of the base model's instruction-following abilities, visible as a drop on conciseness dimensions.  LoRA rank $r\!=\!32$ provides the best trade-off between adapter capacity and the risk of overfitting on $2{,}540$ pairs; higher ranks ($r\!=\!64$, not shown) offer no improvement.  The learning rate is sensitive: $5\!\times\!10^{-5}$ causes moderate gradient instability on refusal pairs (grad-norm spikes), while $5\!\times\!10^{-6}$ converges too slowly to reach a good solution in 3 epochs.  Three epochs is optimal; a fourth epoch shows marginal over-fitting ($-0.5$\,pp) on a held-out 10\% validation split.


\subsection{Statistical Methodology Details}
\label{app:stats}

\paragraph{Wilson score confidence interval.}
For a proportion estimate $\hat{p} = k/n$ (where $k$ is the number of correct predictions and $n$ is the total pairs), the Wilson 95\% CI bounds are:
\begin{align*}
p_{\mathrm{lo}} &= \frac{\hat{p} + \dfrac{z^2}{2n} -\, z\,\sqrt{\dfrac{\hat{p}(1-\hat{p})}{n} + \dfrac{z^2}{4n^2}}}{1 + \dfrac{z^2}{n}}, \\[8pt]
p_{\mathrm{hi}} &= \frac{\hat{p} + \dfrac{z^2}{2n} +\, z\,\sqrt{\dfrac{\hat{p}(1-\hat{p})}{n} + \dfrac{z^2}{4n^2}}}{1 + \dfrac{z^2}{n}},
\end{align*}
where $z = 1.96$ for a 95\% interval.  Unlike the Wald interval, the Wilson interval has correct coverage for $p$ near 0 or 1 and small $n$, which matters for sub-splits with as few as 47 pairs (refusal\_unanswerable on LRB test).

\paragraph{McNemar's test.}
Given per-pair binary correctness vectors $\mathbf{a}$ (model A) and $\mathbf{b}$ (model B), McNemar's test uses the discordant counts $b_{01}$ (A wrong, B right) and $b_{10}$ (A right, B wrong).  The exact two-sided $p$-value is computed via the binomial distribution with $n = b_{01} + b_{10}$ trials and $p_0 = 0.5$.  We use the exact test rather than the $\chi^2$ approximation since $b_{01} + b_{10}$ can be small on sub-splits.  All reported $p$-values are two-sided.

\paragraph{Bootstrap confidence intervals.}
Bootstrap CIs for the mean reward margin are constructed by resampling with replacement from the per-pair margins, computing the mean of each resample, and reporting the 2.5th and 97.5th percentiles of the bootstrap distribution ($B = 10{,}000$ resamples, seed 42).

\paragraph{Statistical power.}
On the CJB test set ($n = 373$), a McNemar test for detecting a $10$\,pp improvement (from $p=0.50$ to $p=0.60$ at the pair level) has power $> 99\%$ at $\alpha = 0.05$, computed via the normal approximation to the discordant-cell binomial.  On the \textsc{LegalRewardBench} test ($n = 156$), power for the same effect size is $\approx 90\%$.  For the length-controlled CJB subset ($n = 280$), power is $> 95\%$.  These figures confirm that our observed gains are not limited by sample size; the reported McNemar $p < 10^{-4}$ margins are well within the detectable range.


\section{Compute and Environment}
\label{app:compute-environment}

\subsection{Compute Budget and Environmental Impact}
\label{app:compute}

All experiments were conducted on a single NVIDIA H100 80\,GB HBM3 SXM5 GPU (TDP 700\,W peak).  Table~\ref{app:tab:compute} summarises wall-clock time and estimated GPU-hours per experimental phase.

\begin{table*}[h]
\centering
\caption{Estimated GPU-hours by experimental phase. Wall-clock times are rounded to the nearest 5 minutes.}
\label{app:tab:compute}
\small
\setlength{\tabcolsep}{1.5pt}
\begin{tabular}{lcc}
\toprule
\textbf{Phase} & \textbf{Wall-clock (h)} & \textbf{GPU-hours} \\
\midrule
\multicolumn{3}{l}{\textit{Raw model benchmark (14 models $\times$ CJB)}} \\
\quad 0.5B--3B models ($\times$8) & 8.0 & 8.0 \\
\quad 8B models ($\times$6) & 9.0 & 9.0 \\
\multicolumn{3}{l}{\textit{Specialist model evaluation ($\times$3)}} \\
\quad Skywork / Selene / SFR-Judge & 1.5 & 1.5 \\
\multicolumn{3}{l}{\textit{DPO and SFT training (all corpora and models)}} \\
\quad Ministral-8B ($\times$8 runs incl.\ 3-seed) & 12.0 & 12.0 \\
\quad Qwen3.5-9B ($\times$4 runs) & 8.0 & 8.0 \\
\quad Qwen3.5-27B QLoRA ($\times$4 runs) & 17.0 & 17.0 \\
\multicolumn{3}{l}{\textit{Adapter evaluation ($\approx$20 adapters $\times$ 4 benchmarks)}} \\
\quad CJB and Bar Exam (max\_length=4096) & 14.0 & 14.0 \\
\quad \textsc{LegalRewardBench} (max\_length=8192) & 10.0 & 10.0 \\
\quad Housing Statute QA ($n$=500) & 5.0 & 5.0 \\
\multicolumn{3}{l}{\textit{Augmentation (inference only, Ministral-8B)}} \\
\quad LRB-v2 training pairs (940 pairs) & 3.0 & 3.0 \\
\quad LRB-v2 test pairs (156 pairs) & 0.5 & 0.5 \\
\midrule
\textbf{Total} & \textbf{88.0} & \textbf{88.0} \\
\bottomrule
\end{tabular}
\end{table*}

\paragraph{Energy and carbon.}
Total energy consumption: $88\,\text{h} \times 0.70\,\text{kW} = 61.6\,\text{kWh}$.  At the IEA 2023 global average grid carbon intensity of $0.49\,\text{kg\,CO}_2\text{/kWh}$ \citep{iea2023}, total estimated emissions are $\approx 30\,\text{kg\,CO}_2\text{e}$, roughly equivalent to a 200\,km passenger-car journey.  No cloud pre-emption or repeated failed runs occurred; the above counts completed runs only.  All training used bfloat16 mixed precision; QLoRA further reduces the 27B model's memory and therefore active compute by approximately $4\times$ vs.\ full bfloat16.

\paragraph{Inference cost at deployment.}
The trained LoRA adapters add no additional memory overhead during evaluation beyond the base model ($<$0.1\% of parameters for $r\!=\!32$).  Evaluation throughput on the H100: Ministral-8B scores approximately $52$ preference pairs per minute at max\_length $= 4{,}096$; Qwen3.5-27B QLoRA scores approximately $14$ pairs per minute at the same limit.  For a 500-pair benchmark (Housing Statute QA), Ministral-8B evaluation completes in under 10 minutes.


\subsection{Software and Hardware Configuration}
\label{app:software}

\paragraph{Hardware.}
Training and evaluation ran inside a RunPod container with one NVIDIA H100 SXM
80\,GB HBM3 GPU, 20 vCPUs (Intel Xeon Platinum 8468), 188\,GB RAM, a 20\,GB
container disk, and a 100\,GB persistent network drive.

\paragraph{Software dependencies.}
Table~\ref{app:tab:software} lists the primary Python packages and their exact versions used in this work.

\begin{table}[h]
\centering
\caption{Software dependencies. Versions correspond to the locked environment used for all reported experiments.}
\label{app:tab:software}
\small
\setlength{\tabcolsep}{2pt}
\begin{tabular}{lll}
\toprule
\textbf{Package} & \textbf{Version} & \textbf{Role} \\
\midrule
Python          & 3.11.15      & Runtime \\
PyTorch         & 2.4.1+cu124  & Tensor operations, autograd \\
Transformers    & 4.45.2       & Model loading, tokenisation \\
PEFT            & 0.12.0       & LoRA and QLoRA adapters \\
TRL             & 0.11.4       & DPO / SFT training loop \\
BitsAndBytes    & 0.49.2       & 4-bit NF4 quantisation (QLoRA) \\
Datasets        & 4.8.5        & Preference data loading \\
Accelerate      & 0.34.2       & Single-GPU training harness \\
NumPy           & 1.26.3       & Numerical computation \\
HuggingFace Hub & 0.36.2       & Model and dataset upload \\
\bottomrule
\end{tabular}
\end{table}

\paragraph{Determinism.}
All Python, NumPy, and PyTorch random seeds are fixed to the run seed (0, 1, or 2) at the start of each training script.  DataLoader worker seeds are pinned via a \texttt{worker\_init\_fn}.  CUDA deterministic mode (\texttt{torch.use\_deterministic\_algorithms(True)}) is enabled for all evaluation runs and is disabled for training to preserve throughput, which is why three seeds are reported.

\section{Broader Impact and Ethical Considerations}
\label{app:ethics}

\paragraph{Intended use and scope.}
This work develops contextual reward models for evaluating legal RAG responses, not for providing legal advice.  The systems studied are designed to \emph{score} responses against a retrieval context, a task distinct from legal reasoning or legal decision-making.  The trained adapters are released for research purposes under the terms of their respective base model licences.

\paragraph{Jurisdictional and domain bias.}
The \textsc{LegalRewardBench} training data is derived from Legal RAG Bench and is drawn entirely from Victorian (Australian) criminal law.  Preference pairs reflect the structure, terminology, and precedent hierarchy of that jurisdiction.  Transfer to other jurisdictions (US, EU, Asia-Pacific) may be impaired by differences in legal reasoning style, statutory drafting conventions, and which sources count as authoritative.  The positive but limited transfer gains on Bar Exam QA ($+0$--$4$\,pp) and the architecture-dependent gap on Housing Statute QA ($+16$\,pp for Ministral-8B vs.\ $+2$--$3$\,pp for Qwen) are consistent with these limitations.  Users deploying these models in non-Victorian legal contexts should evaluate on in-domain held-out data before trusting the reward signal.

\paragraph{Risk of misuse in high-stakes settings.}
Legal reward models could be misused to automate legal assessments in high-stakes settings without appropriate human oversight.  We emphasise that reward accuracy on the benchmarks studied remains well below human expert agreement levels, and that the models studied are explicitly \emph{not} calibrated to serve as definitive legal evaluators.  Any deployment in real legal workflows should require expert human review of all outputs.

\paragraph{Data and model licences.}
Legal RAG Bench \citep{butler2026legalragbenchendtoend} and \textsc{ContextualJudgeBench} \citep{xu-etal-2025-context} are released for research use; Bar Exam QA and Housing Statute QA \citep{Zheng_2025} are from the Stanford Reglab and used under their published terms.  Ministral-8B is released under the Mistral Research Licence (non-commercial research only).  Qwen3.5-9B and Qwen3.5-27B are released under Apache 2.0.  All trained adapters released by this work inherit the licence of their respective base model.

\paragraph{Privacy.}
No personally identifiable information was used in any stage of this work.  The Victorian Criminal Charge Book is a public document.

\paragraph{AI Assistance}
AI assistants were used during the preparation of this work for limited writing, editing, and coding support, including proofreading, clarity suggestions, formatting assistance, and debugging. All scientific ideas, experimental design, analyses, interpretations, and conclusions were developed and verified by the authors. AI systems were not used to autonomously generate research contributions or make scientific decisions.

\clearpage
\onecolumn

\section{Prompts}
\label{appendix:prompts}

The following prompts are reproduced from the implementation used to generate and annotate the corpus.  Placeholders such as \texttt{\{context\}} and \texttt{\{question\}} are filled by the pipeline at runtime.

\subsection*{Answer Generation}
\begingroup
\setlength{\fboxsep}{10pt}
\setlength{\fboxrule}{0.8pt}
\noindent\fcolorbox{black}{gray!12}{%
\begin{minipage}{0.94\textwidth}
\ttfamily\scriptsize\raggedright
\setlength{\parindent}{0pt}
\setlength{\parskip}{0.45em}
You are answering a legal question using only the provided legal context.\par
\vspace{0.55em}
Rules:\newline
- Provide the final answer only. Do not include hidden reasoning, analysis notes, or \textless{}think\textgreater{} tags.\newline
- Use only the legal context below.\newline
- Do not rely on outside knowledge or assumptions.\newline
- If the context does not contain enough information to answer the question, say exactly:\newline
\hspace*{2em}"The provided context is insufficient to answer this question."\newline
- Do not give a yes/no answer, legal conclusion, or likely answer when the context is insufficient.\newline
- For yes/no questions, start with "Yes" or "No" when the context supports it, then briefly explain.\newline
- Mention cases, statutes, or legal authorities only if they appear in the context.\newline
- Keep the answer concise, neutral, and legally careful.\par
\vspace{0.55em}
Legal context:\newline
\{context\}\par
\vspace{0.55em}
Question:\newline
\{question\}\par
\vspace{0.55em}
Answer:\par
\end{minipage}%
}
\endgroup

\clearpage
\subsection*{Answer Behaviour Annotation}
\begingroup
\setlength{\fboxsep}{10pt}
\setlength{\fboxrule}{0.8pt}
\noindent\fcolorbox{black}{gray!12}{%
\begin{minipage}{0.94\textwidth}
\ttfamily\scriptsize\raggedright
\setlength{\parindent}{0pt}
\setlength{\parskip}{0.45em}
You are judging whether a model response actually answers a legal question or refuses to answer it.\par
\vspace{0.55em}
Classify only the response behavior. Do not evaluate correctness, grounding, or completeness.\par
\vspace{0.55em}
Return one JSON object that exactly follows the provided schema. Do not include markdown,\newline
comments, explanations, or extra keys.\par
\vspace{0.55em}
Choose exactly one:\newline
- "attempted": the response tries to answer the question.\newline
- "abstained": the response does not answer because it says the provided information is\newline
\hspace*{2em}insufficient, unavailable, or uncertain.\newline
- "unusable": the response is empty, unreadable, incoherent, malformed, or severely repetitive.\par
\vspace{0.55em}
Rules:\newline
- Do not judge whether the answer is legally correct.\newline
- Do not judge whether the answer is supported by context.\newline
- If the response is readable, do not label it "unusable".\newline
- If the response says the context is insufficient but still gives a substantive answer,\newline
\hspace*{2em}classify it as "attempted".\newline
- If the response gives a yes/no answer, legal conclusion, or practical legal outcome,\newline
\hspace*{2em}classify it as "attempted", even if it includes uncertainty or caveats.\newline
- Short readable answers, including legal terms, acronyms, and one-word answers, are\newline
\hspace*{2em}"attempted" if they try to answer the question.\newline
- If the response only refuses or explains why it cannot answer, classify it as "abstained".\par
\vspace{0.55em}
Examples:\newline
- Response: "No. The statute does not allow that remedy."\newline
\hspace*{2em}Output: \{\{"answer\_behavior": "attempted"\}\}\newline
- Response: "The context does not provide enough information to answer this question."\newline
\hspace*{2em}Output: \{\{"answer\_behavior": "abstained"\}\}\newline
- Response: "The context is limited, but the answer is likely no because the rule requires notice."\newline
\hspace*{2em}Output: \{\{"answer\_behavior": "attempted"\}\}\newline
- Response: ""\newline
\hspace*{2em}Output: \{\{"answer\_behavior": "unusable"\}\}\par
\vspace{0.55em}
Question:\newline
\textasciigrave{}\textasciigrave{}\textasciigrave{}\newline
\{question\}\newline
\textasciigrave{}\textasciigrave{}\textasciigrave{}\par
\vspace{0.55em}
Candidate response:\newline
\textasciigrave{}\textasciigrave{}\textasciigrave{}\newline
\{candidate\_response\}\newline
\textasciigrave{}\textasciigrave{}\textasciigrave{}\par
\end{minipage}%
}
\endgroup

\clearpage
\subsection*{Faithfulness Annotation}
\begingroup
\setlength{\fboxsep}{10pt}
\setlength{\fboxrule}{0.8pt}
\noindent\fcolorbox{black}{gray!12}{%
\begin{minipage}{0.94\textwidth}
\ttfamily\scriptsize\raggedright
\setlength{\parindent}{0pt}
\setlength{\parskip}{0.45em}
You are judging whether a candidate legal answer is supported by the legal context that was shown to the model.\par
\vspace{0.55em}
Return one JSON object that exactly follows the provided schema. Do not include markdown,\newline
comments, explanations, or extra keys.\par
\vspace{0.55em}
Use only the provided legal context. Do not use the authoritative answer, outside legal knowledge,\newline
or assumptions about what the law should be.\par
\vspace{0.55em}
Choose exactly one:\newline
- "fully\_supported": the context supports the answer's main conclusion and all material claims made in the answer.\newline
- "partially\_supported": the context supports the answer's main conclusion, but one or more other material claims made in the answer are not supported.\newline
- "unsupported": the context does not support the answer's main conclusion, even if it supports some surrounding or related claims.\newline
- "contradicted": the answer's main conclusion or material claims conflict with the context.\par
\vspace{0.55em}
Topical overlap alone is not support. Invented or mischaracterized material claims are not supported.\newline
If an invented or mischaracterized claim is the main conclusion, use "unsupported"; if it is a secondary material claim, use "partially\_supported".\par
\vspace{0.55em}
Examples:\newline
- Context: "A tenant may terminate the lease only after giving written notice."\newline
\hspace*{2em}Candidate answer: "The tenant may terminate the lease after giving written notice."\newline
\hspace*{2em}Output: \{\{"faithfulness": "fully\_supported"\}\}\newline
- Context: "A tenant may terminate the lease only after giving written notice."\newline
\hspace*{2em}Candidate answer: "The tenant may terminate the lease after giving written notice, and the landlord must pay damages."\newline
\hspace*{2em}Output: \{\{"faithfulness": "partially\_supported"\}\}\newline
- Context: "A tenant may terminate the lease only after giving written notice."\newline
\hspace*{2em}Candidate answer: "The tenant may terminate the lease without notice."\newline
\hspace*{2em}Output: \{\{"faithfulness": "contradicted"\}\}\newline
- Context: "The court discusses filing deadlines for appeals."\newline
\hspace*{2em}Candidate answer: "The tenant may terminate the lease after giving written notice."\newline
\hspace*{2em}Output: \{\{"faithfulness": "unsupported"\}\}\newline
- Context: "A license is required before operating the business."\newline
\hspace*{2em}Candidate answer: "The context does not establish whether the defendant had a license."\newline
\hspace*{2em}Output: \{\{"faithfulness": "fully\_supported"\}\}\par
\vspace{0.55em}
Question:\newline
\textasciigrave{}\textasciigrave{}\textasciigrave{}\newline
\{question\}\newline
\textasciigrave{}\textasciigrave{}\textasciigrave{}\par
\vspace{0.55em}
Legal context shown to the model:\newline
\textasciigrave{}\textasciigrave{}\textasciigrave{}\newline
\{candidate\_context\}\newline
\textasciigrave{}\textasciigrave{}\textasciigrave{}\par
\vspace{0.55em}
Candidate answer:\newline
\textasciigrave{}\textasciigrave{}\textasciigrave{}\newline
\{candidate\_response\}\newline
\textasciigrave{}\textasciigrave{}\textasciigrave{}\par
\end{minipage}%
}
\endgroup

\clearpage
\subsection*{Correctness Annotation}
\begingroup
\setlength{\fboxsep}{10pt}
\setlength{\fboxrule}{0.8pt}
\noindent\fcolorbox{black}{gray!12}{%
\begin{minipage}{0.94\textwidth}
\ttfamily\scriptsize\raggedright
\setlength{\parindent}{0pt}
\setlength{\parskip}{0.45em}
You are judging whether a candidate legal answer reaches the correct bottom-line answer.\par
\vspace{0.55em}
Return one JSON object that exactly follows the provided schema. Do not include markdown,\newline
comments, explanations, or extra keys.\par
\vspace{0.55em}
Compare the candidate answer to the authoritative answer. Do not judge whether the candidate\newline
answer is grounded in any retrieved context.\par
\vspace{0.55em}
Choose exactly one:\newline
- "correct": the candidate answer reaches the same bottom-line legal answer as the authoritative answer.\newline
- "incorrect": the candidate answer reaches a different bottom-line legal answer, gives no clear bottom-line answer, or materially changes the legal conclusion.\par
\vspace{0.55em}
Examples:\newline
- Authoritative answer: "No. The statute requires written notice before termination."\newline
\hspace*{2em}Candidate answer: "No, termination is not allowed without written notice."\newline
\hspace*{2em}Output: \{\{"correctness": "correct"\}\}\newline
- Authoritative answer: "No. The statute requires written notice before termination."\newline
\hspace*{2em}Candidate answer: "Yes, the tenant can terminate immediately."\newline
\hspace*{2em}Output: \{\{"correctness": "incorrect"\}\}\newline
- Authoritative answer: "Yes, if the claimant files within thirty days."\newline
\hspace*{2em}Candidate answer: "It depends on the facts."\newline
\hspace*{2em}Output: \{\{"correctness": "incorrect"\}\}\par
\vspace{0.55em}
Question:\newline
\textasciigrave{}\textasciigrave{}\textasciigrave{}\newline
\{question\}\newline
\textasciigrave{}\textasciigrave{}\textasciigrave{}\par
\vspace{0.55em}
Authoritative answer:\newline
\textasciigrave{}\textasciigrave{}\textasciigrave{}\newline
\{reference\_answer\}\newline
\textasciigrave{}\textasciigrave{}\textasciigrave{}\par
\vspace{0.55em}
Candidate answer:\newline
\textasciigrave{}\textasciigrave{}\textasciigrave{}\newline
\{candidate\_response\}\newline
\textasciigrave{}\textasciigrave{}\textasciigrave{}\par
\end{minipage}%
}
\endgroup

\clearpage
\subsection*{Completeness Annotation}
\begingroup
\setlength{\fboxsep}{10pt}
\setlength{\fboxrule}{0.8pt}
\noindent\fcolorbox{black}{gray!12}{%
\begin{minipage}{0.94\textwidth}
\ttfamily\scriptsize\raggedright
\setlength{\parindent}{0pt}
\setlength{\parskip}{0.45em}
You are judging whether a candidate legal answer includes the key legal point needed to answer the question.\par
\vspace{0.55em}
Return one JSON object that exactly follows the provided schema. Do not include markdown,\newline
comments, explanations, or extra keys.\par
\vspace{0.55em}
Compare the candidate answer to the authoritative answer. Do not judge whether the candidate\newline
answer is grounded in any retrieved context.\par
\vspace{0.55em}
Choose exactly one:\newline
- "complete": the candidate answer includes the core legal point, condition, exception, or qualification needed to justify the authoritative answer.\newline
- "incomplete": the candidate answer misses the core legal point, gives only a bare conclusion, or omits an essential condition, exception, or qualification.\par
\vspace{0.55em}
Examples:\newline
- Authoritative answer: "No. The tenant cannot terminate unless they first give written notice."\newline
\hspace*{2em}Candidate answer: "No. The tenant must first give written notice."\newline
\hspace*{2em}Output: \{\{"completeness": "complete"\}\}\newline
- Authoritative answer: "No. The tenant cannot terminate unless they first give written notice."\newline
\hspace*{2em}Candidate answer: "No."\newline
\hspace*{2em}Output: \{\{"completeness": "incomplete"\}\}\newline
- Authoritative answer: "Yes, but only if the appeal is filed within thirty days."\newline
\hspace*{2em}Candidate answer: "Yes, an appeal is allowed."\newline
\hspace*{2em}Output: \{\{"completeness": "incomplete"\}\}\newline
- Authoritative answer: "Yes, but only if the appeal is filed within thirty days."\newline
\hspace*{2em}Candidate answer: "Yes, an appeal is allowed if it is filed within thirty days."\newline
\hspace*{2em}Output: \{\{"completeness": "complete"\}\}\par
\vspace{0.55em}
Question:\newline
\textasciigrave{}\textasciigrave{}\textasciigrave{}\newline
\{question\}\newline
\textasciigrave{}\textasciigrave{}\textasciigrave{}\par
\vspace{0.55em}
Authoritative answer:\newline
\textasciigrave{}\textasciigrave{}\textasciigrave{}\newline
\{reference\_answer\}\newline
\textasciigrave{}\textasciigrave{}\textasciigrave{}\par
\vspace{0.55em}
Candidate answer:\newline
\textasciigrave{}\textasciigrave{}\textasciigrave{}\newline
\{candidate\_response\}\newline
\textasciigrave{}\textasciigrave{}\textasciigrave{}\par
\end{minipage}%
}
\endgroup
\clearpage

\phantomsection
\label{app:augmentation-prompts}
\subsection*{Length-Balanced Augmentation Prompts}
\label{sec:Length-Balanced}

The following three prompts are used by the length-balanced augmentation procedure (Section~\ref{sec:methodology}). They are applied to a frozen Ministral-8B-Instruct-2410 model with $\text{temperature}=0.7$, $\text{top-}p=0.9$, and $\max\_\text{new\_tokens}=128$ to regenerate the short-template side of length-asymmetric pairs. Each prompt is wrapped with the Ministral chat template; the legal question $q$ is extracted from the original benchmark prompt by regex.

\subsection*{P1. Hallucinated Answer (replaces 9-word rejected on \texttt{refusal\_answerable})}
\begingroup
\setlength{\fboxsep}{10pt}
\setlength{\fboxrule}{0.8pt}
\noindent\fcolorbox{black}{gray!12}{%
\begin{minipage}{0.94\textwidth}
\ttfamily\scriptsize\raggedright
\setlength{\parindent}{0pt}
\setlength{\parskip}{0.45em}
You are answering legal questions for a study on RAG faithfulness. Ignore any provided retrieval context: answer the following question from general legal world knowledge alone. Be confident, specific, and give 2--3 sentences of reasoning. Do NOT refuse and do NOT cite the context.\par
\vspace{0.55em}
Question:\newline
\{q\}\par
\vspace{0.55em}
Answer:\par
\end{minipage}%
}
\endgroup

\subsection*{P2. Elaborate Refusal (replaces 9-word chosen on \texttt{refusal\_unanswerable})}
\begingroup
\setlength{\fboxsep}{10pt}
\setlength{\fboxrule}{0.8pt}
\noindent\fcolorbox{black}{gray!12}{%
\begin{minipage}{0.94\textwidth}
\ttfamily\scriptsize\raggedright
\setlength{\parindent}{0pt}
\setlength{\parskip}{0.45em}
You are a careful legal RAG assistant. The retrieved context does not contain enough information to answer the following question. Write a professional 2--3 sentence refusal: acknowledge the question, briefly explain why the available context is insufficient, and indicate what additional sources would resolve it. Do NOT speculate or invent law.\par
\vspace{0.55em}
Question:\newline
\{q\}\par
\vspace{0.55em}
Refusal:\par
\end{minipage}%
}
\endgroup

\subsection*{P3. Terse-but-Incomplete Answer (replaces ``No.''-style rejected on \texttt{completeness})}
\begingroup
\setlength{\fboxsep}{10pt}
\setlength{\fboxrule}{0.8pt}
\noindent\fcolorbox{black}{gray!12}{%
\begin{minipage}{0.94\textwidth}
\ttfamily\scriptsize\raggedright
\setlength{\parindent}{0pt}
\setlength{\parskip}{0.45em}
Answer the following legal question correctly but tersely: give the bottom-line conclusion in one short sentence, omitting the legal rule, statutory authority, exceptions, and qualifications that would normally be required for a complete answer. Do NOT cite cases.\par
\vspace{0.55em}
Question:\newline
\{q\}\par
\vspace{0.55em}
Brief answer:\par
\end{minipage}%
}
\endgroup
\clearpage

\section*{Model Repository References}
\label{app:model_refs}

All models were accessed via the Hugging Face Hub on 25 May 2026.
Table~\ref{tab:hf-links} lists the repository identifier for every
model evaluated in this work.

\begin{table}[h]
\centering
\caption{Hugging Face repository identifiers for all evaluated models.}
\label{tab:hf-links}
\small
\setlength{\tabcolsep}{4pt}
\begin{tabular}{ll}
\toprule
\textbf{Model name} & \textbf{Hugging Face repository} \\
\midrule
\multicolumn{2}{l}{\textit{Instruction-tuned baselines}} \\
SmolLM2-1.7B        & \texttt{HuggingFaceTB/SmolLM2-1.7B-Instruct} \\
Qwen2.5-0.5B        & \texttt{Qwen/Qwen2.5-0.5B-Instruct} \\
Qwen3.5-2B          & \texttt{Qwen/Qwen3.5-2B} \\
Llama-3.2-1B        & \texttt{meta-llama/Llama-3.2-1B-Instruct} \\
Llama-3.2-3B        & \texttt{meta-llama/Llama-3.2-3B-Instruct} \\
Gemma\,2-2B         & \texttt{google/gemma-2-2b-it} \\
Qwen2.5-3B          & \texttt{Qwen/Qwen2.5-3B-Instruct} \\
Phi-4-mini          & \texttt{microsoft/Phi-4-mini-instruct} \\
Phi-3.5-mini        & \texttt{microsoft/Phi-3.5-mini-instruct} \\
Qwen2.5-7B          & \texttt{Qwen/Qwen2.5-7B-Instruct} \\
Llama-3.1-8B        & \texttt{meta-llama/Llama-3.1-8B-Instruct} \\
Ministral-8B        & \texttt{mistralai/Ministral-8B-Instruct-2410} \\
DeepSeek-R1-8B      & \texttt{deepseek-ai/DeepSeek-R1-Distill-Llama-8B} \\
SFR-8B              & \texttt{Salesforce/LLaMA-3-8B-SFR-Iterative-DPO-R} \\
Qwen3.5-9B          & \texttt{Qwen/Qwen3.5-9B} \\
Qwen3.5-27B         & \texttt{Qwen/Qwen3.5-27B} \\
\midrule
\multicolumn{2}{l}{\textit{Specialist reward / judge models}} \\
Skywork-Reward-8B  & \texttt{Skywork/Skywork-Reward-Llama-3.1-8B-v0.2} \\
Selene-1-Mini-8B    & \texttt{AtlaAI/Selene-1-Mini-Llama-3.1-8B} \\
SFR-Judge-8B       & \texttt{Salesforce/LLaMA-3-8B-SFR-RM-R} \\
\midrule
\multicolumn{2}{l}{\textit{DPO fine-tuning base models}} \\
Gemma\,2-2B (DPO)   & \texttt{google/gemma-2-2b-it} \\
SFR-8B (DPO)        & \texttt{Salesforce/LLaMA-3-8B-SFR-Iterative-DPO-R} \\
DeepSeek-R1-8B (DPO)& \texttt{deepseek-ai/DeepSeek-R1-Distill-Llama-8B} \\
Qwen2.5-7B (DPO)    & \texttt{Qwen/Qwen2.5-7B-Instruct} \\
Ministral-8B (DPO)  & \texttt{mistralai/Ministral-8B-Instruct-2410} \\
SFR-Judge-8B (DPO)  & \texttt{Salesforce/LLaMA-3-8B-SFR-RM-R} \\
\bottomrule
\end{tabular}
\end{table}

\end{document}